\documentclass[aps,prl,onecolumn,superscriptaddress,groupedaddress]{revtex4}
\usepackage{subfig}
\usepackage{graphicx}  
\usepackage{dcolumn}   
\usepackage{bm}        
\usepackage{amssymb}   
\usepackage{slashed}
\usepackage{graphicx}				
\usepackage{amsmath}
\usepackage{mathtools}
\usepackage{tikz,pgf}
\usepackage{comment}
\usepackage{asymptote}
\usetikzlibrary{arrows,backgrounds}
\usetikzlibrary{fit,scopes,calc,matrix,positioning,decorations.pathmorphing}
\usepackage[all]{xy}
\usepackage{yfonts}
\newcommand{\bra}[1]{\ensuremath{\left\langle#1\right|}}
\newcommand{\ket}[1]{\ensuremath{\left|#1\right\rangle}}
\newcommand{\Bracket}[1]{\ensuremath{\left\langle#1\right\rangle}}

\begin{document}
\title{Experimental verification of the quantum nature of a neural network}
\author{Andrei T. Patrascu}
\address{FAST Foundation, Destin FL, 32541, USA\\
email: andrei.patrascu.11@alumni.ucl.ac.uk}
\begin{abstract}
Neural networks are being used to improve the probing of the state spaces of many particle systems as approximations to wavefunctions and in order to avoid the recurring sign problem of quantum monte-carlo. One may ask whether the usual quasi-classical neural networks have some actual hidden quantum properties that make them such suitable tools for a highly coupled quantum problem. I discuss here what makes a system quantum and to what extent we can interpret a neural network as having quantum remnants. I suggest that a system can be quantum both due to its fundamental quantum constituents and due to its dynamics (seen as maps in a category), therefore, we can obtain entanglement both due to the quantum constituents' nature and due to the functioning rules, or, in category theory terms, both due to the quantum nature of the objects of a category and of the maps. From a practical point of view, I suggest a possible experiment that could extract entanglement from the dynamics (maps) of an otherwise cvasi-classical (from the point of view of the constituents) neural network. 
\end{abstract}
\maketitle
\section{Introduction}
Neural networks are nowadays regarded as computational tools that have a broad set of applications from quantum chemistry to solving logistics problems. However, in all biological circumstances, neural networks appear as special materials that have a flexible design reaching down to the cellular level and with an impact on various other tissues like muscles, vascular systems, etc. As such, a neural network is a special material with surprising physical, chemical, and, as I wish to present in this article, quantum properties. 

The goal of this article is to show that a quasi-classical neural network may have underlying quantum properties that can be amplified by its dynamics. It must be noted that the neural network referred to in this article is not a numerical simulation, as a simulation would ultimately, by design, have no complex phase and hence would not present quantum effects. My discussion is based on a real physically implemented quasi-classical neural network. The main observation is that such a neural network may generate quantum effects due to the dynamics (or, in a categorical sense, due to the maps between the objects, not the objects themselves). The end result is that a quasi-classical neural network, due to its dynamics, could generate entangled (quantum) states and could be used for entanglement generation and extraction. As such, this is of particular interest to subjects like quantum biology or in describing quantum phenomena in natural neural networks where the usual criteria of temperature and scale would otherwise block such effects.

Probably the best way of starting this article is to ask a question: what does it mean to be quantum? 
This question is particularly important because systems or mathematical structures that we may usually associate to quasi-classical physics may have unexpected quantum properties. From another point of view, systems usually simulated by means of strictly classical rules may, upon physical implementation, present quantum behaviour that escaped our previous analysis. Quantum mechanics has several defining principles, two of the most important being: its probabilistic nature: the outcomes generated by quantum mechanical calculations are always probabilistic, even if that probability gets very close to 0 or 1, and second, there is no pre-determined uniquely defined state of a system in the absence of an observable capable of asking a question about that property of the system. This property makes the probability amplitude and hence the wavefunction of the system represent the maximal knowledge attainable about the system in a given context.
The wavefunction, being subject to a linear differential equation, allows the usage of the principle of superposition to find new solutions, with the consequence that physical systems, described by such probability amplitudes, can exist in what is known as superposed states. Moreover, the quantitative properties of a state of a system are generally not uniquely defined, and sometimes are not even defined at all in the absence of an observable to be associated to a method of determining them. Not any observable is capable of fully determining the numerical value of a property of the state of a system, and some observables may not be compatible with respect to a given property. As a result, some of the matrix valued observables of quantum mechanics may not commute with each other. 
Due to these properties, the wavefunction is, somehow unfortunately, called "non-realistic", and that is unfortunate because there is nothing "unreal" about the wavefunction, in fact it provides the maximum of information about the system, and in many cases far more than a classical system could ever hope to achieve. A wavefunction ought to be interpreted as a catalogue of maximal knowledge that can be obtained about a property of the system to be analysed. As a consequence, the rules of obtaining the probabilistic outcomes of quantum mechanics change with respect to a classical system. We therefore have to employ a method of obtaining a module of a complex number, from what is a fundamentally complex quantity, namely the wavefunction. This wavefunction needs to be complex in order to account for correlations in our knowledge about the system. In turn, by this method we will generate the well known interference patterns in the probabilistic outcomes of quantum mechanics, by which quantum mechanics differs from classical mechanics. 
That last property of quantum mechanics is Born's rule.

One of the main distinctions between classical and quantum mechanics is therefore that in classical mechanics probabilities are associated strictly to lack of knowledge about otherwise well expressed and physically realised properties of the system, while in quantum mechanics, probabilities emerge out of a lack of knowledge that is inherent to the system, and for which there is no physically realised property of the system to talk of. 

The distinction between these two situations is rather formal, as, of course, nobody could in principle measure a system as being in a superposed state a-priori. To even realise the existence of such superposed states, one would have to perform a statistical experiment and see how the distributions differs from that of a system that we would imagine to satisfy the rules of classical probabilities. Indeed, such statistical experiments have been performed and they overwhelmingly prove the validity of quantum mechanics and the existence of a wavefunction based description of nature. 

This however leaves open the possibility that systems that we imagined to be classical are in fact only quasi-classical and their quantum properties are somehow suppressed in the specific context we constructed for them. This would also rise the question of whether it would be possible by some means, to amplify such quantum properties in otherwise quasi-classical systems so that they can perform certain tasks (like, for example entangling some states) that could become relevant in quantum computers.
This observation is not new. Indeed [37] shows that large structures regain their quantum properties and act as quantum objects in a double slit experiment, given a suitable parametric setup of the experiment.

There exists a somehow widespread assumption that quantum phenomena must be either very small, or happen at low enough energy, or be small corrections of the order of $O(\hbar)$ to classical phenomena and that there is some form of "gap" between a so called "quantum world" and the "real" or "every-day" world.
Moreover, one usually assumes that the limit taken to obtain the classical limit of a quantum system involves trivially taking $\hbar\rightarrow 0$ as is commonly described using the Ehrenfest approximation to calculate expectation values. My thesis is that these assumptions are wrong. It is well known that, by Ehrenfest theorem, the relation to classical mechanics is trivially obtained only when the geometry of the system allows for a well localised wavefunction, which is not the case in most situations of experimental relevance in quantum mechanics. Moreover, the situation changes dramatically when the classical limit involves non-linear dynamics as is the case in quasi-classical physically realised neural networks. In those cases, the relation between quantum dynamics and their quasi-classical limit is in fact much more complicated and interesting. 

The approaches to quantum computing discussed up to now are based on controlling the interference of the "catalogue of knowledge" wave patterns (basically, a sort of wavefunction engineering). Those methods are extremely susceptible to quantum decoherence, namely extended interaction and entanglement with the environment which leads to an exponential increase of the number of terms arising in the "catalog of knowledge" describing the combined system of environment and computational device
If one approaches quantum phenomena from a categorical perspective, one notices that the maps between quantum objects may as well have quantum properties, may become non-cartesian, and can produce and exploit entanglement. The maps between quantum objects are associated to the specific dynamics through which the system goes, and this very dynamics can produce, enhance, and utilise entanglement. 

In particular the non-linear dynamics associated to neural networks appears to have an interesting effect on the underlying quantum properties. One may think that I am discussing two different types of dynamics in an equivalent way, namely on one side the classical dynamics of a neural network, where the non-linearities occur, and on the other side, a quantum dynamics of a quasi-classical neural network seen as some limit of a pure quantum situation that behaves linearly according to the laws of quantum mechanics. While this may appear so if one thinks of nature as being strictly divided between a classical and a quantum realm, my discussion is based on the assumption that such a distinction is not physical, and that in reality all systems have an underlying quantum behaviour that may, in some cases, be to some extent suppressed. The process of suppression is sometimes amenable to a description by Ehrenfest' theorem and by taking trivial limits when $\hbar\rightarrow 0$, but in most cases of interest in the dynamics of neural network, it is not.

One direction of research nowadays is to find some classical underpinnings for quantum mechanics. This research is an important part of the so called "foundational" research on quantum mechanics. While the common approach is to lift or extend the observables of a system from the uniquely defined classical real numbers to matrix valued observables that encode several possible outcomes without the possibility of predicting which one will occur from the data about the system itself, some quantum foundational research tries to find a classical (therefore well determined and unique) way of characterising the system based on some form of hidden classical variables. 
My approach is the exact opposite of this path. Instead of searching for some underpinning classical structure to quantum mechanics, my view is that even systems that appear as classical, may still have hidden quantum properties, due, for example to the quantum nature of the transformation maps between the objects of the specific category considered. 
In particular, systems assumed to be classical, like a neural network, despite being simulated as classical objects in conventional computers, when physically realised, may exhibit quantum phenomena as quasi-classical systems. In particular, the non-linear dynamics of neural networks may in fact amplify quantum behaviour. 

The same procedure that provides us with a quantum structure over the space of objects can provide us with a quantum structure over the space of maps between those objects in a categorical sense. If, for example, we employ geometric quantisation, following the prescriptions of pre-quantisation, polarisation or splitting, and half-form correction, we obtain (in most cases) a well defined structure on the objects (states). We could perform the same on the morphisms (or maps) of the category and obtain a quantum structure for those. In fact, it turns out that quantum information carried on the object space is found also in the space of morphisms, and sometimes quantum properties of morphisms (like the cartesian non-separability leading to entanglement) may appear in the space of morphisms and have an effect on the space of objects as well.

There are several macroscopic systems that are generally assumed to be classical, that by the way in which they work, have structures that give them some quantum properties, as far as the definitions above are concerned. One of these systems is the neural network. It is important to mention that there have been essays in which one speculated about the possibility of generating quantum computing processes in big enough biological neural networks by focusing on some detailed mechanism in some molecular substructure of their neural network [1]. I think such an approach is not feasible on a large scale (although one has discovered entanglement of spins in some animal brains [2,3] and substantial progress is being done in understanding macroscopic quantum phenomena from this point of view). Those methods are extremely susceptible to decoherence. However, one may ask what if the large structure of a network has properties that at least in part reflect the fundamental axioms of quantum mechanics?

In this case, the entanglement between spins in animal brains would not be occurring, as described until now, in a substance assimilated to a heat bath, but in an environment adapted to amplify and preserve quantum superpositions by means of the non-linear dynamics resulting from the neural properties it possesses. In this case it would be less surprising to see biological brains taking advantage of quantum superpositions and act like a quantum annealer.

There have also been essays in constructing so called quantum neural networks [4]. Those networks have the links represented by quantum processes implemented by unitary transformations, representing a quantum neural network in the form of a large quantum circuit. Those methods are promising and it would be amazing if they will be achieved. However, there is still the problem of decoherence that will require such networks to be extremely robust, unfortunately a property still beyond today's reach. 

It has however been noticed that classical neural networks do a great job at solving many body quantum problems with strong coupling [5], that are fundamentally quantum and expected to be properly solved on quantum computers.
One role neural networks play in solving strongly correlated quantum systems is through simplifying the wavefunction by providing us with a more compact representation thereof. 
Another role played by neural networks is to improve the understanding of quantum dynamics [12-16]. 
What are the properties of a classical neural network that would make it suitable for solving such complex quantum problems? While it is true that classical neural networks seem to be very suitable for simulating quantum processes, I wish to ask in this article whether this is due to the fact that the rules of functioning of a neural network that appears classical from the perspective of its objects (namely neurons) has in fact hidden quantum properties at the level of the morphisms (maps) between those objects. Also, it would be interesting to see how the non-linear behaviour of the quasi-classical neural dynamics can affect underlying quantum degrees of freedom.

\section{a brief categorical interpretation}
The following brief categorical digression is here to underline what it means to be quantum, something that has indeed surprised early physicists but that should be quite obvious to us now. 
The mathematical underpinnings of such definitions are axiomatisation [6], which allows us to obtain the most abstract form of a concept, that can then be used to find it in various other forms and circumstances, usually not too obvious to a non-axiomatic approach, and categorification [7], that allows us to expand certain objects or collections of objects into categories by adding the required maps (morphisms) between the respective objects. Categories are very versatile, giving us the possibility of apparently working in completely unrelated fields, but which have the map and object structures acting equivalently or in a way that can be related to some underlying common category. If we try to do this with quantum mechanics we may notice that many structures, mathematical or computational, may have some level of quantumness, associated to some or all of the axioms, while being applied in completely different areas and seeming to a non-categorial mind as purely classical. 

Indeed, various quantum properties appear to be general properties not only of physical particles, but also of a series of mathematical structures that have a broad applicability. Non-separability, as encountered in the definition of entanglement, is also found, under the guise of non-cartesianity in the construction of fibre bundles, in which basically the bundle measures the global failure of maintaining the local cartesianity of a certain pairing. A non-trivial fibre bundle is defined to characterise precisely such a failure of the cartesian product between the fibre and the base to expand globally. This is just a mathematical example that could make the reasoning I try to follow here clearer: quantum properties are not bound to particles, but can also characterise various rules by which dynamical systems function. Such quantum properties make them function differently compared to their classical counterparts. 
Of course, this doesn't mean a classical fibre bundle is quantum. We need additional structure for this. We need some type of quantum polarisation (like in geometric quantisation) and some form of half form correction, both general properties of quantisation. However, there exists a deep similarity between the category of cobordisms whose objects are (n-1)-dimensional manifolds and its morphisms are n-dimensional cobordisms, and the catgeory of Hilb whose objects are Hilbert spaces used to describe states and the morphisms are linear operators describing "processes" [32]. With this relation, the cartesian non-separability property that leads to entanglement in quantum mechanics has a similar role in other situations that involve non-separability, as is the case in fibre bundles.  

From a practical perspective, it is important to know that the same type of entanglement exists in a category related by a functor to the initial one, and it would be possible to harness it as a quantum resource and use it in circumstances that are apparently remote from the initial category. 
In particular, an apparently classical neural network may extract its entanglement from an associated category and behave like a quantum system when its dynamics is analysed.
A wavefunction, or a probability amplitude can be associated not only to the objects of the category, but also to the maps between those objects, and hence superposition and even entanglement described at the level of the wavefunction associated to the quantum maps may prove to be a new quantum resource that could be harnessed by physical neural networks. 
This would lead to the possibility of having quantum computation like effects occurring even in the most noisy and hot environments, like the ones existing in animal brains.
A physical feature emerging from the quantum properties above is the inseparability of state spaces of quantum states [8]. This obstruction to cartesian pairing has been traditionally called entanglement, and it is advisable to keep that notion even when applied to more abstract mathematical objects [9], [32].

Entanglement is considered a characterising feature of quantum mechanics. It is a clearly non-classical phenomenon capable of offering correlations above any that could be obtained classically [10]. Entanglement is based on the non-separability of state spaces of systems that are connected to form an over-arching super-system. The non-separability relies on two concepts. First, if we take two systems, each having their state spaces given, and combine them one obtains a paired state space with a dimension larger than the sum of the dimensions of the two subspaces. The dimension will grow as the product of the dimensions of the two subspaces, leading to a pairing that involves significant global information that cannot be mapped or retrieved by local measurements on any of the two subsystems separately. The fact that pairing is not always cartesian is a feature of quantum mechanics that has correspondences in many algebraic structures. One can link this with the commutation relations that become non-abelian for incompatible observables, and finally with a probabilistic interpretation that allows, by means of statistics, to determine interference patterns linked to the global structure of the manifold on which the phenomena occur. In all cases, the two main aspects of the "quantumness" of a system are the linear product structure, and its ability to provide access to non-local features by means of a probabilistic, wave-function based interpretation. It is worth noting that the entangled states are, as any quantum state, maximal states of knowledge. They are being defined to describe global properties of the combined system in the case in which separated independent descriptions of each of the subsystems that are combined are unachievable and don't even exist for the specified observables in an unambiguous sense. 

We usually regard entanglement as a quantum phenomenon, and rightfully so. Mathematically however, entanglement is an obstruction to cartesianity in the sense that, as opposed to the $Set$ category, describing sets and having as maps functions linking elements in the set, the $Hilb$ category doesn't allow for simple cartesian splitting. In that sense, if we bring together two systems, each described by their respective space of states, we obtain a space of states larger than the cartesian pairing between the two. Therefore the combined system will have potential states that cannot be found in the separate initial spaces of states of the two individual systems. Those entangled states are a specificity of the $Hilb$ category which is a non-cartesian monodromic category. However, there are other non-cartesian monodromic categories, like for example the category of cobordisms, and we can describe functors relating those two categories, making the physics of cobordisms equivalent up to a point with quantum physics [32]. 

In this article I show that a (quasi)-classical neural network can indeed be represented by a non-cartesian monodromic category and that there exists a functor that links it to the $Hilb$ category. By doing so, a normal (quasi)-classical neural network will show properties of quantum entanglement at a large scale in the same way in which other quantum systems show [37]. However, being highly decoherent at the level of the constituent objects, it will be up to the maps to act as a resource for quantum entanglement. This means that the quantum nature emerges from the rules we impose on how the network should work, not from the actual quantum nature of its constituents, but finally, there should be no difference between a "natural" quantum system made out of fundamental constituents that are "quantum", and one that is forced to function by rules (maps or morphisms of the category) that result in an overall quantum behaviour of the network. The parameters of the network can be encoded in a Hilbert Kernel, where the respective parameters play the role of linear superpositions of the learned results, similar to the quantum superpositions allowed by the complex phases of a wavefunction. Therefore, there appears to be a strong analogy (if not more, a functorial connection) between entangled quantum states and neural networks and between the category of Hilbert spaces and the supposed category of neural networks.

Indeed it may be possible to experimentally test the morphisms origin of some quantum behaviour of a neural network by looking at a physical implementation of a neural network that appears to be classical from the perspective of the objects involved, but may gain quantum properties due to its inherent morphisms. Such physical neural networks have been constructed [17]. Until such an experiment can be performed, I can offer a numerical simulation showing that indeed this is the case, namely that provided with proper input, a physical, classical neural network can entangle the particles that it receives as input.

\section{how is a classical neural network quantum?}
A (quasi)-classical neural network has certain properties that could be thought of as quantum to some extent. While we do not propagate a quantum state through a series of quantum gates forming a hard to maintain network, we instead use a set of quasi-classical neurons that, acting together, would be able to produce a state that could be classified as quantum for all practical purposes.

Let us consider a neural network, made out of neurons firing according to classical rules. Let us also consider several layers of such neurons, as can be imagined in a deep learning network. The parameters of the neurons are bound together in a series of linear combinations in which the activation function plays the role of a non-linear contributor to the adaptability of the network. The relation among the neurons however is always linear. A signal is transmitted to the network as an input, which is passed then through the network and a gradient descent backpropagation method is used to train the linearly connected neurons forming the network to the desired structure. After defining a loss function, backpropagation is required to transmit the end-point error back to the input node and to optimise the parameters in the process. This process allows the network to access, via its non-linear activation function and particularly through its loss function, the non-local structure of the features it explores. This is also why neural networks are so good at classifying non-local classification problems and are finally used as approximations for highly entangled quantum states. But are classical neural networks just approximations of quantum systems, or are they, in some sense, quantum themselves, not through the interactions that make them function, but through their global structure and their backpropagation and loss function features (mainly their dynamics maps)? Usually, in a neural network, the optimised result is not easily associated with the input, which is why artificial neural networks have the well known problem of inferring causal connections. This may not be surprising given that the end-result is obtained through a process of extensive linear superposition of learned data and then optimised via a loss function that introduces the learned data to the global structure of the problem it is asked to solve. This is what an entangling gate would do. 

A Hadamard gate would create a linear superposition of the initial state. This is what occurs to our initial data via the network linear superposition. A CNOT gate would combine two states, switching the result of one when the other is in a specified state. This is generally the result of our loss functions. The only difference is that we are not a-priori acting on quantum states but on suitably represented initial learning data. It all depends on what meaning we give to that data and how we interpret the output then, to make it a quantum analogue or a simple classical superposition. 

In nature, biological neural networks work through the process of accepting signals in the form of collections of molecular ions that reach the synapse passing through the pre-synaptic region. They are not unique signals, but statistical distributions of such molecular ions that present strong correlations with each other due to the mechanisms that produced them in the pre-synaptic region. 
Therefore the real situation in nature is much more complicated than what the current simplified models present us. Treating the input as a well define classical signal is certainly an over-simplification. The signal in fact results as an emergent property of a distribution of molecular ions. Whether the distribution has or has not quantum correlation is a question not completely answered. In particular, the whole process of transmitting the distribution of molecular ions in the synaptic region, receiving it by the receptors of the neuron, and then triggering a series of mechanisms inside the neuron could be a quasi-classical process, but how exactly the quasi-classical dynamics is obtained as a limit of quantum phenomenon is not understood as for now. 

As I discussed in the previous chapter, taking the classical limit $\hbar\rightarrow 0$ is not trivial and in most cases the non-linear classical dynamics in a neural network would create quite interesting problems. 
In order to describe such processes, I bring in an example of working with the Wigner-Weyl representation. The main goal of this approach is to find a structure that comes closest to a quantum distribution over the phase space. This is of course technically impossible due to the non-commutativity of momentum and position in quantum mechanics. If we want to calculate expectation values, which is the case for example in the applications of Ehrenfest's theorem, we need some form of distributions. This can be achieved by considering Wigner distributions and Weyl transformed operators. Indeed, given a density matrix 
\begin{equation}
\hat{\rho}=\ket{\psi}\bra{\psi}
\end{equation}
we combine the process of having a Weyl transformed operator 
\begin{equation}
\tilde{A}(x,p)=\int e^{-i p y/\hbar}\bra{x+y/2}\hat{A}\ket{x-y/2}dy
\end{equation}
and the formation of a Wigner distribution starting from the expectation value of the density matrix 
\begin{equation}
\bra{x}\hat{\rho}\ket{x'}=\psi(x)\psi^{*}(x')
\end{equation}
in the form 
\begin{equation}
W(x,p)=\tilde{\rho}/h=\frac{1}{h}\int e^{-ipy/\hbar}\psi(x+y/2)\psi^{*
}(x-y/2)dy
\end{equation}
Defining the trace of two operators as
\begin{equation}
Tr[\hat{A}\hat{B}=\frac{1}{h}\int \int \tilde{A}(x,p)\tilde{B}(x,p)dx dp
\end{equation}
we obtain for the expectation value of an operator 
\begin{equation}
\Bracket{A}=\int \int W(x,p)\tilde{A}(x,p)dx dp
\end{equation}
If we want to analyse the dynamics we look at the partial time derivatives of the Wigner distribution
\begin{equation}
\frac{\partial{W}}{\partial t}=\frac{1}{h}\int e^{-ipy/\hbar}[\frac{\partial \psi^{*}(x-y/2)}{\partial t}\psi(x+y/2)+\frac{\partial\psi(x+y/2)}{\partial t}\psi^{*}(x-y/2)]dy
\end{equation}
By looking at the Schrodinger equation 
\begin{equation}
\frac{\partial\psi(x,t)}{\partial t}=-\frac{\hbar}{2im}\frac{\partial^{2}\psi(x,t)}{\partial x^{2}}+\frac{1}{i\hbar}U(x)\psi(x,t)
\end{equation}
we can split the time evolution of the Wigner distribution as in 
\begin{equation}
\frac{\partial W}{\partial t}=\frac{W_{T}}{\partial t}+\frac{\partial W_{U}}{\partial t}
\end{equation}
The kinetic part we are treating easily as 
\begin{equation}
\frac{\partial W_{T}}{\partial t}=\frac{1}{4\pi i m}\int e^{-ipy/\hbar}[\frac{\partial^{2}\psi^{*}(x-y/2)}{\partial x^{2}}\psi(x+y/2)-\psi^{*}(x-y/2)\frac{\partial^{2}\psi(x+y/2)}{\partial x^{2}}]dy
\end{equation}
which can be written as 
\begin{equation}
\frac{\partial W_{T}}{\partial t}=-\frac{p}{m}\frac{\partial W(x,p)}{\partial x}
\end{equation}

and the potential part which can be written as 
\begin{equation}
\frac{\partial W_{U}}{\partial t}=\frac{2\pi}{ih^{2}}\int e^{-ipy/\hbar}[U(x+y/2)-U(x-y/2)]\psi^{*}(x-y/2)\psi(x+y/2)dy
\end{equation}
which by means of a series expansion can be written as 
\begin{equation}
\frac{\partial W_{U}}{\partial t}=\sum_{s=0}(-\hbar^{2})^{s}\frac{1}{(2s+1)!}(\frac{1}{2})^{2s}\frac{\partial^{2s+1}U(x)}{\partial x^{2s+1}}(\frac{\partial}{\partial p})^{2s+1}W(x,p)
\end{equation}
we see however that taking the $\hbar\rightarrow 0$ limit here is not trivial as we also have a $(2s+1)$ derivative of $W(x,p)$ which would always bring us more $\hbar$ to the denominator, than what we could eliminate and hence the $\hbar\rightarrow 0$ limit would be ill-defined. 
Usually to avoid this problem, the transition from quantum to classical is taken by introducing mixed states. Indeed, classical distributions are linear in phase space, while Wigner functions are not. We can write linear superpositions of wavefunctions $\psi=\psi_{\alpha}+\psi_{\beta}$ but from that it doesn't follow the superposition of the associated Wigner functions, therefore we cannot conclude that $W_{\psi}=W_{\alpha}+W_{\beta}$. If we introduce mixed states, however 
\begin{equation}
W(x,p)=\tilde{\rho}/h=\sum_{j}P_{j}W_{j}(x,p)
\end{equation}
we recover the linearity as is the case for classical phase space distributions, but indeed, we use mixed states. The advantage is that, while indeed we use mixed states, we can see easier how the quantum terms appear and in particular we can safely take the $\hbar\rightarrow 0$ limit. 
For the sake of an example, I will use as basic model the harmonic oscillator ground state. In the case we consider pure states, we would have 
\begin{equation}
\psi=A[\psi_{0}(x-b)+\psi_{0}(x+b)]
\end{equation}
and we use the ground state of the harmonic oscillator
\begin{equation}
\psi_{0}(x)=\frac{1}{\pi^{1/4}\sqrt{a}}e^{-x^{2}/(2a^{2})}
\end{equation}
leading to the Wigner distribution 
\begin{equation}
W(x,p)=\frac{1}{h(1+e^{-b^{2}/a^{2}})}e^{-(ap)^{2}/\hbar^{2}}[e^{-(x-b)^{2}/a^{2}}+e^{-(x+b)^{2}/a^{2}}+2e^{-x^{2}/a^{2}}cos92bp/\hbar)]
\end{equation}
In this case we would have a superposition of quantum states.
If we work however with mixed states we write a linear combination of classical states and therefore we obtain 
\begin{equation}
W(x,p)=\frac{1}{2}[W_{0}(x-b,p)+W(x+b,p)]
\end{equation}
where
\begin{equation}
W_{0}(x,p)=\frac{2}{h}exp(-a^{2}p^{2}/\hbar -x^{2}/a^{2})
\end{equation}
and therefore
\begin{equation}
W(x,p)=\frac{1}{h}e^{-a^{2}p^{2}/\hbar^{2}}[e^{-(x-b)^{2}/a^{2}}+e^{-(x+b)^{2}/a^{2}}]
\end{equation}
Applying now the partial derivative $(\frac{\partial}{\partial p})^{2s+1}$ on the Wigner functions of mixed states
\begin{equation}
W(x,p)=\int W_{0}(x,p-p_{0})P(p_{0})dp_{0}
\end{equation}
 defined by means of a probability density 
\begin{equation}
P(p_{0})=\frac{1}{c\sqrt{\pi}}e^{-p_{0}^{2}/c^{2}}
\end{equation}
results in 
\begin{equation}
W(x,p)=\frac{1}{\pi\sqrt{a^{2}c^{2}+\hbar^{2}}}e^{-x^{2}/a^{2}}e^{p^{2}/(c^{2}+\hbar^{2}/a^{2})}
\end{equation}
In general we couldn't fix $p$ or $x$ because of the uncertainty relations of quantum mechanics that links them through $\hbar$. Indeed the dynamics can reverse the places of x and p (we can obtain squeezed states, etc.) therefore fixing one is not realistic. If we fix the width on $x$, then the width on $p$ cannot be arbitrarily small. However, once we shift to mixed states, the width of other $p$-states can be arbitrary (large or small) and therefore we can safely take the limit $\hbar\rightarrow 0$. 
In this way in the example of a double coherent pure state, taking the classical limit by means of mixed states, we obtain 
\begin{equation}
W(x,p)=\frac{1}{2\pi a c(1+e^{-(b/a)^{2}})}e^{-p^{2}/c^{2}}[e^{-(x-b)^{2}/a^{2}}+e^{-(x+b)^{2}/a^{2}}+2e^{-x^{2}/a^{2}}e^{-b^{2}/a^{2}}]
\end{equation}
where the last term expresses the level of suppression of quantum properties in a continuous manner. 
However, non-linear classical dynamics in this approach to transitioning to the classical limit may result in very large higher order derivatives. Combined with our $\hbar\rightarrow 0$ limit, this may indeed amplify quantum effects by means of classical non-linear dynamics. 
As neural networks do implement non-linear classical dynamics, it seems that controlling the non-linearity parameters of a quasi-classical neural network may induce or amplify quantum effects. 

In any sense, neural networks in general are devices capable of linking global and local information by using a method that reminds us of quantum correlations. Consider for example the situation describing reinforced learning. To do that we have to think of one or a set of agents interacting with an environment and developing a strategy resulting in the maximisation of a reward function. The maximisation problem is obviously non-local, and the agents are trained / learn global structures in their early training phase. A policy that could be learned this way would be the conditional flipping of one state when a certain type of input is detected.

One could argue that the backpropagation may be a very artificial construct, specific to simple minded neural networks like those used nowadays to solve various practical tasks. This may be true, but as we can see now, backpropagation is just a tool used to obtain a certain set of extremisations, and extremisation is in principle a non-local problem in the sense that information from various points of the manifold must be included to establish where a certain maximum or minimum (relative or not) lies. An extremum is not a property of a point, but a property of the environment. I will not discuss this further, but for the type of potential arising in neural networks quite generally, even more strongly non-local extremisation methods may be required for proper convergence [31].

\section{where are the wavefunctions?}
A quantum system is described either by a wavefunction (in case we have a pure system) or by a density matrix which unifies the statistical and quantum description. A benchmark of a quantum system is however the existence of a quantum pre-probability, or a complex wavefunction that has the property that its modulus square provides a probability distribution. Such a wavefunction appears as an eigenfunction of some observable matrix operator (particularly the Hamiltonian), associated with a specific eigenvalue (particularly an Energy), and can be propagated by means of a Schrodinger type equation. The question that rises is where does such a structure, or even a bra-ket vector formalism arise in a neural network. Answering this question may reveal the domain in which the neural network behaves more like a quantum system and can be used to provide quantum computational efficiency (in an absolute or approximate way) for neural computational architectures. 
One of the approaches to understanding the quantum nature of neural network, mainly found in ref. [35] (of which I became aware only after finishing the first draft of this article myself) is to show that quantum mechanics is a thermodynamically emergent property of a neural network. It basically claims that the dynamics of a neural network is fundamentally classical, but, due to a thermodynamic behaviour we obtain a set of optimal equations of motion that resemble Schrodinger's equation and provide us with a type of solution that strongly resembles the wavefunction. This however is not a wavefunction as understood in quantum mechanics until a generalisation is performed, through which the neural system is described by a grand-canonical ensemble, in which effort is being made to make the complex phase of the solution multivariate and fundamentally non-determinable. The final result is that this type of indetermination makes the number of neurons non-determinable as well, which results in the "emergence" of a quantum behaviour. In that article, the essence of the appearance of a quantum behaviour is linked to the existence of a grand-canonical ensemble which makes the complex phase of a solution of a Madelung equation multivariate and the number of neurons undetermined. The conclusion was that the neural network will behave quantum if one considers the thermodynamical limit of a large number of neurons and a chemical potential that links the system ensemble with its grand-canonical bath. The number of neurons will therefore become undetermined in the given context and the complex phase of the solution of the Madelung equation would become unobservable. 
This is a very interesting approach, but I consider it limited in various ways, and in fact I do not consider quantum dynamics (and of course the wavefunction) to be emergent in a thermodynamical sense, from a presumably classical "microscopic" neural network dynamics. It is definitely true that making a certain quantity undetermined means that the system will retain quantum properties with respect to that quantity, but I consider the connection to quantum mechanics to be much deeper and with broader implications. The solution of ref. [35] was to make the states (or numbers) of the inner (hidden) neurons unobservable and to claim that the result would be quantum. However, if the neural network has to retain some quantum properties, those must not be related to thermodynamical limits, and in fact, as will be shown here, are not. 
Let us go through two apparently disconnected subjects, one being gauge invariance and the meaning of gauge variant observables, and the other being observables and their determinability in classical and quantum mechanics. 
When we discuss about observables in interacting theories, the main aspect we have to pay attention to is to have those observables gauge invariant. The choice of a gauge is equivalent to the choice of a frame (like a reference frame) on the respective fibres of a fibre bundle. Obviously those can be chosen in any way we desire, and therefore we expect that the observables on which we all can agree upon must be gauge choice invariant. However, if we don't impose these restrictions, we can perfectly well define observables that are not gauge invariant, let's call those gauge variant. A gauge variant observable would be one on which only those observers will agree upon, that made the same choice of a gauge. Experiments can be constructed that provide answers to questions formulated in terms of gauge variant observables. The only difficulty will be that observers making different (and fundamentally arbitrary) gauge choices, will not agree upon the answer. Whether this is a big issue or not, is something to be considered. In any case, there are two interesting situations in which this aspect is relevant: general relativity and Quantum Chromodynamics. We know now, that aside from certain approximations, observables in general relativity cannot be both local and gauge invariant. Observers will only agree upon observables that are fundamentally non-local. The reason for this is the curvature of spacetime which brings in non-local modifications that have to be taken into account when defining an observable that doesn't depend on a choice of a diffeomorphism transformation. Something similar happens in QCD, with the only difference that now, the transformations are not acting on spacetime but instead in a gauge space of the theory. The "curvature" here is given by the field stress tensor $F_{\mu\nu}$ but the rest is similar. We do not have a gauge invariant "localised" observable in this space either (say the space of colour), the only difference, which makes QCD a fully local quantum field theory, is that the gauge invariant non-local observables are not non-local in spacetime but instead in the gauge space. In fact, we may have observers that, given their choice of gauge, will not agree upon the colour of a quark, as various interactions may change it. Any colour based observable will be "local" in the gauge space but will not be gauge invariant, and different observers may detect different colours. This is why colour is not a property that we can detect freely in nature. The observers will always detect "white" states, combinations of either colour and anti-colour or three colours combined. However, in general relativity we also can approximately define local gauge invariant observables, when the curvature is too small to matter at the respective scales. The same thing is valid in QCD where small enough curvature (in the asymptotic region of high energies, as opposed to low energies in general relativity) allows us to detect and be aware of the existence of colour as an extra degree of freedom. Therefore both general relativity and QCD share a similar situation regarding local gauge-invariant observables: they can only be approximate. If we want true gauge invariance, we need to make the observables non-local. There is another aspect to it: to make the theory consistent from the perspective of the wavefunctions (or fields) involved, we must make certain properties undetermined or imprecise, at the level of either the gauge space (colour) or at the level of spacetime (like in localised spacetime positions). 

Now let us see what happens in quantum mechanics: there, observables are hermitian matrices that describe the fact that the outcomes for a certain property of a certain system are not determined by the given experimental setup. While a certain outcome may occur at a given measurement, this outcome is not fully determined, as another observation will likely produce another result for the same quantity.
The quantum state of a system is described by a wavefunction which has a complex phase that describes basically its non-observable nature, particularly what a specific experiment cannot determine all by itself, but that becomes observable only via the repetition of various experiments. This repetition of events is the "limit" that corresponds in general relativity or in QCD to the limit of small curvature (analogously speaking). The multi-valued nature of quantum observables is therefore linked to the existence of the quantum phase of a wavefunction. The propagation of a wavefunction is described by an amplitude, which we may observe in an experiment, and a complex phase, which plays basically the role of creating a consistent description along the propagation of the wavefunction. To introduce it however in one single experimental cycle, this phase must play the role of an arbitrary function, not dissimilar to that we encounter in the construction of gauge theories.

Exactly like in the context of gauge invariance, this complex phase is not determined, but without it, all experimental results related to interference patterns of probability distributions would not be describable. The introduction of gauge invariance in a physical quantum field theory therefore, not surprisingly, is implemented exactly in the phase structure of a complex wavefunction first. Retaining this gauge invariance in order to describe only non-ambiguous observables therefore results in the construction of a covariant derivative in the gauge space and in the introduction of gauge fields (connections). Therefore, both quantum mechanics and gauge theories (describing interactions) require the introduction of an invariance to a certain type of transformation in a non-determinable complex phase, non-determinable in the sense that it cannot be measured directly, but has effects on the construction of measurement statistics or on the fact that interactions become possible. 

The authors of ref. [35] made a similar approach for the case of a neural network described by a grand-canonical ensemble. While they did not notice the gauge structure involved, they also found that the state of a neural network is described by a Hamilton-Jacobi type equation, and constructed a Madelung and Schrodinger type dynamics, with solutions that are looking like our wavefunctions in quantum mechanics. Their derivation was based on the assumption that the neural network is a thermodynamical system and they constructed a new symmetry in the theory that allowed for a complex phase that would render the free energy function multivariate and the number of neurons undetermined. I find this approach very ingenious but still resulting from a somewhat mechanistic and ultimately incorrect view of the process. The problem is not the consistency of the theory as formulated in a thermodynamic sense, but the fact that indeed, quantum mechanics doesn't have to emerge from thermodynamical arguments, not even in the case of a neural network, where its presence is assured by the neural dynamics, independent of us taking a thermodynamical limit.

Instead, I will show that a similar construction appears when considering any form of gauge theory describing a neural network's learning process. Going from the construction of a gauge symmetry in the action functional describing the neural network allows us to obtain wavefunction type behaviour and quantum effects that are not related to any thermodynamical considerations. Instead, the neural network seems to be able to act as a quantum network at any scale, and any number of neurons. 
If we describe the dynamics of a neural network, as shown in [19] by means of a Hamilton Jacobi equation, the existence of a gauge invariance as well as the existence of quantum behaviour will rely on the non-invertibility of the $(q,p)$ equations and ultimately of the canonical transformation generating function $F$. There, the unified view of the gauge arbitrary function and the quantum phase arbitrary function will be made clear.
A gauge theory is one in which the dynamical variables are specified up to transformations that link different arbitrary reference frames. In quantum mechanics, this arbitrariness is related to the choice of transformations of the complex phase, which can be such that they impose the standard gauge invariance, say $U(1)$ or $SU(2)$ or is determined by the fact that there is no pre-defined set of observables that are complete and compatible in quantum mechanics. Both situations result in similar indeterminacies and unobservable shifts or transformations of the complex quantum phase. For a neural network to be described by a wavefunction one needs to have a set of undetermined quantities, and a symmetry transformation on the complex phase, but those do not need to emerge in a thermodynamical sense. Instead they are linked to a gauge invariance that can be assumed to be part of the dynamics of a neural network.

The universe seems to have a strange preference of providing us with physical systems that seem to be described by more variables than there are physically independent degrees of freedom. While we always think "lowly" of these variables, we could hardly do anything without them. We know for a fact that they are not "physical" or "real" but we find them both in gauge field theories and in some interpretation of quantum mechanics (say, hidden variables). They appear to describe what Schrodinger called "states of knowledge" about a system (or wavefunctions in modern language). We also find them in neural networks, as the "hidden" variables or the states of the bulk neurons of a neural network. Those variables are not really there, and theories that try to consider otherwise fail in various well known ways (see for example hidden variables in quantum mechanics). If however we try doing physics without them, we get stuck with the impossibility of describing the basic forms of interactions or relying on differential equations and Cauchy problems to solve problems. Not only are local variables hidden, but so are even topological or global properties. Topologists know only too well that topological properties may or may not be manifest according to the structures we use to detect them. A reference to this idea is for example [36] but there are plenty of other sources as well.

A classical system is described by 
\begin{equation}
S_{L}=\int_{t_{1}}^{t_{2}}L(q,\dot{q})dt
\end{equation}
If we require this to be stationary under variations $\delta q^{n}(t)$ that vanish at the endpoints, we obtain the Euler Lagrange equations
\begin{equation}
\frac{d}{dt}(\frac{\partial L}{\partial \dot{q}^{n}})-\frac{\partial L}{\partial q^{n}}=0
\end{equation}
for $n=1,...,N$. In order to see the type of transformation we require when we impose this stationarity condition we are well advised to rewrite the equation as 
\begin{equation}
\ddot{q}^{n'}\frac{\partial L}{\partial \dot{q}^{n'}\partial \dot{q}^{n}}=\frac{\partial L}{\partial q^{n}}-\dot{q}^{n'}\frac{\partial^{2} L}{q^{n'}\partial\dot{q}^{n}}
\end{equation}
In this equation we can relate the acceleration on the left hand side, with the positions and velocities on the right hand side, if the matrix $\frac{\partial L}{\partial \dot{q}^{n'}\partial \dot{q}^{n}}$ is invertible. In that case, a Cauchy problem is uniquely defined and we can have a single trajectory to follow. 
However, if the determinant of the above matrix is zero, the accelerations will not be uniquely determined by such positions and velocities, and we must introduce also arbitrary functions in our solution. Those arbitrary functions are basically choices of frames that we can make at any point in time and that are indeed arbitrary, as there is no predefined choice for them. Those arbitrary functions will make the evolution of the system undetermined because not all variables will evolve predictably. 
This is the origin of gauge freedom, and, at the same time, the origin of a complex phase in quantum mechanics. In both theories, the same quantities make the system essentially undetermined in its time evolution, despite the fact that the equations of motion in both cases are strictly deterministic. Historically, there is however a distinction on how we deal with those two aspects of physics, but it is my conviction that this distinction is only formal and not fundamental to Nature. 
The existence of arbitrary functions in the solutions of a Cauchy problem means that the system has "gauge degrees of freedom". In gauge theory, invariance of the solutions to such gauge choices makes us able to define interactions unambiguously. In quantum mechanics, the existence of such "phase choices" allows us to take into account more options than what is "physically" there and allows us to define a "state of knowledge" (hence a wavefunction). Entanglement is nothing but the correlation between instances that involve such non-physical properties of intermediate states. Surely, a correlation that involves all intermediate states that could be there but are not formally realised will be much stronger than any correlation that involves only well determined properties of the system. This property is of great importance in performing quantum computations. A correlation between gauge variant states however could prove equally if not more important to performing calculations that are not accessible to classical computers. The fundamental distinction is in my opinion minimal.
Let us continue and define a Hamiltonian formalism in this context. We do that by introducing the momenta
\begin{equation}
p_{n}=\frac{\partial L}{\partial \dot{q}^{n}}
\end{equation}
The existence of arbitrary function solutions is a result of the vanishing of the determinant 
\begin{equation}
det(\frac{\partial^{2}L}{\partial \dot{q}^{n}\partial \dot{q}^{n'}})=0
\end{equation}
This condition leads to the fact that we cannot invert the velocities as functions of $(q,p)$, and therefore 
$v=v(q,p)$ is non-invertible. This means that not all momenta are independent, which leads to a set of primary constraints relating them: 
\begin{equation}
\phi_{m}(q,p)=0
\end{equation}
with $m=1,...,M$. This results in the fact that inverse transformation from $p$ to $\dot{q}$ is indeed multivalued. If we provide a point $(q^{n}, p^{n})$ that solves the constraints, the inverse map $(q^{n},\dot{q}^{n})$ that solves the equation for momenta $p_{n}=\frac{\partial L}{\partial \dot{q}^{n}}$ is not unique. To make the transformation single-valued one needs to introduce several extra parameters that specify the $\dot{q}$ on the target manifold. These parameters are Lagrange multipliers. 
We already see that gauge theories and quantum mechanics seem to have a common origin. In both cases, either the state of knowledge (wavefunction) in quantum mechanics, or the solution to an equation of motion, gains arbitrary functions added to the solution, making them capable of encoding unobserved information that describes either interaction (gauge theory) or quantum information. This discussion may also be relevant for the age-old question on the origin of gauge symmetry, and could provide an alternative (quantum) explanation to the origin of the standard model group $U(1)\times SU(2)\times SU(3)$. 
The constraint surface can be represented in various ways but in order to be able to construct a functional Hamiltonian formalism there are some restrictions on the primary constraints that need to be imposed. They are called "regularity conditions". The constraints $\phi_{m}(q,p)=0$ define a submanifold of the phase space. The vanishing determinant condition brings in a rank $N-M'$ leaving $M'$ independent constraint equations. This makes the primary constraint surface to be of dimension $2N-M'$. The momentum equation $p_{n}=\frac{\partial L}{\partial \dot{q}^{n}}$ defines a map from a $2N$ dimensional space of positions and velocities, to a $2N-M'$ dimensional manifold of the constraints. The inverse images of a point of the constraint manifold therefore won't be single valued. 
The constraint manifold should therefore be covered by open regions where the constraint functions can be split into independent constraints $\phi_{m'}=0$, $m'=1,...,M'$ defined by the condition that the Jacobian matrix 
\begin{equation}
\frac{(\partial \phi_{m'})}{\partial q_{n},p_{n}}
\end{equation}
will be of rank $M'$ on the constraint surface, and dependent constraints $\phi_{\bar{m}'}=0$, $\bar{m'}=M'+1,...,M$ which hold as a result of the other constraints. 
This is equivalent with saying either that the functions $\phi_{m'}$ can be locally taken as the first $M'$ coordinates of a new coordinate system in the vicinity of the constraint surface (definition of a new reference frame), or that the gradients $d\phi_{1},...,d\phi_{M'}$ are locally linearly independent on the constraints surface, or that the variations $\delta \phi_{m'}$ are of order $\epsilon$ for arbitrary variations $\delta q^{i}$ and $\delta p^{i}$ of order $\epsilon$. 
These conditions result in the following results: first, if a smooth phase space function $G$ vanishes on the surface $\phi_{m}=0$ then $G=g^{m}\phi_{m}$ for some functions $g^{m}$. Then, if $\lambda_{n}\delta q^{n} +\mu^{n}\delta p_{n}=0$ for arbitrary variations tangent to the constraint surface $\delta q^{n}$, $\delta p_{n}$ then
\begin{equation}
\begin{array}{c}
\lambda_{n}=u^{m}\frac{\partial \phi_{m}}{\partial q^{n}}
\\
\mu^{n}=u^{m}\frac{\partial \phi_{m}}{\partial p_{n}}
\end{array}
\end{equation}
for some $u_{m}$. 
We may have redundant constraints which make the functions $u_{m}$ non-unique. The Hamiltonian then can be defined in a canonical way as
\begin{equation}
H=\dot{q}^{n}p_{n}-L
\end{equation}
The velocities however enter only in the form of combinations $p(q,\dot{q})$ defined by $p_{n}=\frac{\partial L}{\partial \dot{q}^{n}}$. This results as a property of the Legendre transformation. 
This Hamiltonian however is not uniquely determined as a function of $p$ and $q$. We know this because $\delta p_{n}$ are restricted by the constraint equations $\phi_{m}=0$. The canonical Hamiltonian is only defined on the submanifold of the primary constraints. Outside that surface it can be extended arbitrarily for example as
\begin{equation}
H\rightarrow H+c^{m}(q,p)\phi_{m}
\end{equation}
If we calculate the variations of the hamiltonian 
\begin{equation}
\begin{array}{c}
\delta H =\dot{q}^{n}\delta p_{n} + \delta \dot{q} p_{n}-\delta \dot{q}^{n}\frac{\partial L}{\partial \dot{q}^{n}}-\delta q^{n}\frac{\partial L}{\partial q^{n}}=\\
=\dot{q}^{n}\delta p_{n}-\delta q^{n}\frac{\partial L}{\delta q^{n}}
\end{array}
\end{equation}
which we can re-write as 
\begin{equation}
(\frac{\partial H}{\partial q^{n}}+\frac{\partial L}{\partial q^{n}})\delta q^{n}+(\frac{\partial H}{\partial p_{n}}-\dot{q}^{n})\delta p_{n}
\end{equation}
leading to 
\begin{equation}
\begin{array}{c}
\dot{q}^{n}=\frac{\partial H}{\partial p_{n}}+u^{m}\frac{\partial \phi_{m}}{\partial p_{n}}\\
\\
-\frac{\partial L}{\partial q^{n}}|_{\dot{q}}=\frac{\partial H}{\partial q^{n}}|_{p}+u^{m}\frac{\partial \phi_{m}}{\partial q^{n}}
\end{array}
\end{equation}
This is the standard way in which gauge theories recover the velocities $\dot{q}^{n}$ from the knowledge of the momenta, despite the fact that the determinant was initially null and hence arbitrary functions emerged in the solutions. The gauge theoretical solution to the problem is the introduction of a set of extra parameters $u^{m}$ which determined the position on the larger manifold of the respective objects on the smaller manifold. What is essentially a one-to-many map that is non-invertible and multivalued, becomes a one to one map onto a manifold in which the constraints lead to a series of extra parameters that uniquely determine the solution. 
The deep question that can be asked now is how "fundamental" are those extra parameters, and the answer is : not at all. How is that similar to quantum mechanics? To understand this let us describe a simple experiment that is the foundation of understanding entanglement in quantum mechanics. 
Let us consider two electrons of spin $1/2$ that appeared due to the disintegration of a state of spin $0$. 
At this point we know that the two electrons will have opposite spin, and we can propagate their evolution according to this constraint. However, the knowledge that the spins of the electrons are opposite is incomplete, because, in a sense, the question was ill formulated. When I asked "what is the spin of the electrons", and I answered "it should be $1/2$ and opposite on the two electrons" I made two mistakes. First the spin projection can be opposite, not the spin itself, which remains $1/2$. The second mistake: I did never provide an axis on which the projection is supposed to be defined. The initial, spin $0$ state did not provide such an axis, being a symmetric system. The experiment, as I described it above, also did not provide such an axis, and therefore, the state of the system doesn't have a defined axis, therefore, all the available information is that the spin projections should be opposite, no matter what axis one chooses. Therefore, the spin projection, no matter what axis, must be described by an evolution equation, of which solution, as a solution of the equation of motion, will be, as in the development above, dependent on an arbitrary function, namely the choice of an axis. We can add the extra parameters, those are precisely the axes on the two electrons, say, we choose the axes on the two parameters along our $z$ axis and we measure the spin projections on our new axes. We find the spin projections to be as expected $+\frac{1}{2}$ and $-\frac{1}{2}$. What we can conclude is that given the new parameters of the theory, the axes, the resulting projections are as measured. However, the experiment repeats itself as we wish to provide a statistics of measurements. We decide next time to choose another axis, say the $x$ axis on both electrons. We will again measure and find the spin projections to be $+\frac{1}{2}$ and $-\frac{1}{2}$. We can conclude that given the new choice of parameters, our outcomes are as expected. The novelty is that given the information that the spin projections must be overall opposite, if we measure on one axis, say $z$, the spin projection to be $+\frac{1}{2}$ then we know, without measuring, that the spin will be opposite on a parallel $z$ axis on the other electron. This makes it relatively clear that the additional parameters were not physically realised in any sense, but must be there in order to carry the various options we may choose (a frame, if you wish) at a later moment in time. This is an arbitrary function which we need to propagate alongside the solution of our equation for the description to be meaningful, but that totally depends on choices of "frames" we can make at any different point of time in a different manner. 
In general, in gauge theories, such parameters are seen as "arbitrary choices of coordinates" in the respective positions or local submanifolds, that make the one-to-many map a nice one-to-one map that is "well defined" according to our "sensibilities". But there is no reason not to accept a map that is fundamentally one to many, if the information that has to be carried is in fact of the form "one-to-many". This is what quantum mechanics does, and exactly the same thing is what gauge theory does, the only difference is that in gauge theory, the choices of coordinates and the extra parameters we add in order to make the function one-to-one are less "poignant" than in quantum mechanics. We totally can accept an undetermined set of coordinates, but it is usually less intuitive to accept an undetermined set of properties or features of a system like an electron, atom, the position of an electron in an atom, etc. However, the idea is basically the same. Gauge theories are quantum theories. Historically, gauge theories have been invented early on, and that kind of freedom was not well regarded. People wished to get rid of it, hence they tried to determine the one-to-many map exactly by means of the extra parameters encoding coordinate choices. This worked fine until quantum gauge fields appeared, where more care had to be taken to deal with various sets of one-to-many maps. One had to fix a one-to-many map towards only one outcome, hence fixing the gauge, while allowing for gauge freedom to choose the respective local frames as before, to make interactions possible. In any sense, quantisation of gauge theories is a well known subject, what I am doing here is to provide a more unified view on the matter. 
At this point, we have a set of constraints equations. If they are independent, so will be the vectors 
$\frac{\partial \phi_{m}}{\partial p_{n}}$ on $\phi_{m}=0$. Therefore the $u$s will uniquely determine now the velocities. We can find those $u$s then as functions of the coordinates and velocities by solving 
\begin{equation}
\dot{q}^{n}=\frac{\partial H}{\partial p_{n}}(q,p(q,\dot{q}))+u^{m}(q,\dot{q})\frac{\partial \phi_{m}}{\partial p_{n}}(q,p(q,\dot{q}))
\end{equation}
The Legendre transform from $(q, \dot{q})$ to the constraint surface $\phi_{m}(q,p)=0$ of the new $(q,p,u)$ space being 
\begin{equation}
\begin{array}{c}
q^{n}=q^{n}\\
p_{n}=\frac{\partial L}{\partial \dot{q}^{n}}(q,\dot{q})\\
u^{m}=u^{m}(q,\dot{q})\\
\end{array}
\end{equation}
allows us to define that transformation between spaces of the same dimension and making the map invertible as one obtains the equations
\begin{equation}
\begin{array}{c}
q^{n}=q^{n}\\
\dot{q}^{n}=\frac{\partial H}{\partial p_{n}}+u^{m}\frac{\partial \phi_{m}}{\partial p_{n}}\\
\phi_{m}(q,p)=0\\
\end{array}
\end{equation}
Therefore, we may obtain invertibility of the Legendre transformation at the expense of introducing additional variables. That is usually not what we need to do in quantum mechanics, unless we fully specify the evolution and the experiment demands us to do so. In quantum mechanics we should be very happy with such multivalued expressions as we in fact do transport arbitrary functions (choosing for example the projection axes) across the experimental setup. 
Therefore, if we look back at reference [35] and at their indeterminacy of the number of neurons, such indeterminacy has basically been introduced in the form of a gauge symmetry in the problem, with the goal of allowing arbitrary functions to evolve alongside their solutions of the equations of motion. They definitely did not need thermodynamics for that. The only reason why thermodynamics was introduced was because there was a certain subjective expectation of things "emerging" via thermodynamics, but in reality, I will do the same thing here, keeping other variables as "hidden", that do not need a thermodynamic expression. By making those variables "physically non-realised" I will obtain similar equations, without the need of a thermodynamic approach. 
Without underlining too much the distinctions between my approach and that of [35], it is worth mentioning that [35] was somehow limited in its formulation by a series of assumptions that I consider unphysical: first, that quantum mechanics needs to emerge from a classical sub-structure, and second, that there is a material substructure that is required for a symmetry (be it even gauge symmetry) to exist. In reality, such a material substructure is not needed, everything happening only at the level of potentialities and pretty much nothing at the level of actual physical, measurable realisation. The entanglement that I am showing here to exist in a neural network may as well be between states where correlation is between gauge orbits or between unrealised quantum intermediate states. In both cases, the correlations will be stronger than those we can classically expect, and they are parts of the neural network, even if the inner workings of a neural network are basically classical. But enough with philosophical considerations.
The Hamiltonian equations can also be derived from the variational principle:
\begin{equation}
\delta \int_{t_{1}}^{t_{2}}(\dot{q}^{n}p_{n}-H-u^{m}\phi_{m})dt=0
\end{equation}
for arbitrary variations $\delta q^{n}$, $\delta p_{n}$, $\delta u_{m}$ with the constraints 
\begin{equation}
\delta q^{n}(t_{1})=\delta q^{n}(t_{2})=0
\end{equation}
with the variables $u^{m}$ introduced to make the Legendre transformation invertible. These variables appear as Lagrange multipliers enforcing the primary constraints $\phi(q,p)=0$. The theory is invariant to $H\rightarrow H + c^{m}\phi_{m}$, since this change results in a redefinition of the Lagrange multipliers. The equations of motion can be written as 
\begin{equation}
\dot{F}=[F,H]+u^{m}[F,\phi_{m}]
\end{equation}
where $F(q,p)$ is an arbitrary function of the canonical variables and 
\begin{equation}
[F,G]=\frac{\partial F}{\partial q^{i}}\frac{\partial G}{\partial p_{i}}-\frac{\partial F}{\partial p^{i}}\frac{\partial G}{\partial q_{i}}
\end{equation}
Of course, while the basic principle of quantum mechanics and of gauge theory is similar, the two ideas are somewhat different. The difference lies in the general assumptions we made. In gauge theory, we assume that in principle, we could eliminate gauge symmetry altogether by solving the constraints and re-write the theory as
\begin{equation}
\delta \int_{t_{1}}^{t_{2}}(\dot{q}^{n}p_{n}-H)dt=0
\end{equation}
for independent variations of the coordinates and momenta subject to the constraints $\phi_{m}=0$ and $\delta\phi_{m}=0$. But for this to be true, we still need the regularity conditions for the constraints. 
To see how quantum mechanics is in essence a constraint gauge system we can look at the Hamilton Jacobi equations again, but this time from the perspective of constraints systems. Let's first have a look at the Hamilton Jacobi equation in unconstraint systems. The equations of motion implement a canonical transformation relating the coordinates $q^{i}$ and the momenta $p_{i}$ at each time $t$ and the initial coordinates $q_{0}^{i}$ and momenta $p_{i}^{0}$ at time $t_{0}$ on the other hand. A time independent canonical transformation will link $(q_{0}^{i}, p_{i}^{0})$ to the canonical coordinates $(\alpha^{i},\beta_{i})$. 
A canonical transformation will also link $(q^{i}, p_{i})\rightarrow (\alpha^{i},\beta_{i})$. The generating function is $S=S(q^{i},\alpha^{i},t)$ and we obtain 
\begin{equation}
\begin{array}{c}
p_{i}=\frac{\partial S}{\partial q^{i}}\\
\\
\beta_{j}=-\frac{\partial S}{\partial \alpha^{j}}\\
\end{array}
\end{equation}
and 
\begin{equation}
det(\frac{\partial^{2}S}{\partial \alpha^{j}\partial q^{i}})\neq 0
\end{equation}
The variables $\alpha^{i}$ and $\beta_{i}$ are constants of the motion. Therefore the Hamiltonian describing the dynamics of the evolution of $\alpha^{i}$ and $\beta_{i}$ can be taken to be zero. 
This would be 
\begin{equation}
\bar{H}(\alpha^{i},\beta_{i})=H+\frac{\partial S}{\partial t}=0
\end{equation}
and that produces the Hamilton Jacobi equation
\begin{equation}
\frac{\partial S}{\partial t}+H(q^{i},\frac{\partial S}{\partial q^{i}})=0
\end{equation}
A solution $S(q_{i},\alpha_{i})$ depending on all $n$ variables $\alpha^{i}$ that satisfies the non-zero condition for the determinant above is called a complete integral. If a complete integral is known one can construct a general solution of the equations of motion by simply 
\begin{equation}
\beta_{j}=-\frac{\partial S}{\partial \alpha^{j}}
\end{equation}
The solution of the Hamilton Jacobi equation for which the time-independent canonical variables $(\alpha^{i},\beta_{j})$ are identical with the initial condition $(q_{0}^{i},p_{j}^{0})$ is denoted by $W(q^{i},q_{0}^{i},t)$ and is the Hamiltonian principal function. This will be the classical action 
\begin{equation}
W(q^{i},q_{0}^{i},t)=\int_{(q_{0}^{i},t_{0})}^{(q^{i},t)}du\; L(q,\dot{q})
\end{equation}
However, we also have solutions that depend on fewer integration constants. We call those solutions "incomplete integrals" and we obtain them by fixing $m$ of the $\alpha$ in the complete integral. The unspecified $\alpha$s we call $\alpha^{A}$, $A=1,...,n-m$ and $\alpha_{a}$ those that are fixed. For example we can set all the fixed constants to zero. With this condition, the dependence of $S$ on $\alpha_{a}$ disappears which makes the conjugate variable $\beta^{a}=-\frac{\partial S}{\partial \alpha_{a}}$ undetermined/unknown. This means that the equations 
\begin{equation}
\begin{array}{c}
p_{i}=\frac{\partial S}{\partial q^{i}}\\
\beta_{A}=-\frac{\partial S}{\partial \alpha^{A}}\\
rank(\frac{\partial^{2} S}{\partial \alpha^{A}\partial q^{i}})=n-m\\
\end{array}
\end{equation}
with $\alpha^{A}$ and $\beta_{A}$ given, a complete integral $S(q^{i},\alpha^{A},t)$ can no longer determine a unique solution $(q^{i}(t),p_{i}(t))$ of the equations of motion. Therefore the constants of motion $\alpha^{A}$, $\beta_{A}$ and $\alpha_{a}=0$ do not determine a single classical solution but instead they describe all trajectories that have the same values for the $\alpha^{A}$, $\beta_{A}$ and $\alpha_{a}=0$ but differ in the value of the unknown conjugate $\beta^{a}$. A general situation would then be one in which if we have a solution $(q^{i},p_{i})$ of the above equations at time $t$, and $(q^{i}+\delta q^{i}, p_{i}+\delta p_{i})$ another solution at a later time, then we have
\begin{equation}
\begin{array}{c}
\delta q^{i} = \frac{\partial H}{\partial p_{i}}(q,p)\delta t
\delta p_{i}=-\frac{\partial H}{\partial q^{i}}(q,p)\delta t
\end{array}
\end{equation}
only in the case in which $(q^{i}+\delta q^{i},p_{i}+\delta p_{i})$ and $(q^{i},p_{i})$ are both giving the same values of the conjugate momenta $\beta^{a}$ and hence lie on the same classical trajectory. This happens for the Hamilton Jacobi theory for a complete integral. However, it is also possible to have different values of $\beta^{a}$ for two solutions at different times. Since the variations in $\beta$ are generated by the conjugate variables $\alpha_{a}$ one obtains
\begin{equation}
\begin{array}{c}
\delta q^{i}=[\frac{\partial H}{\partial p_{i}}(q,p)+\lambda^{a}\frac{\partial \alpha_{a}}{\partial p_{i}}(q,p)]\delta t\\
\\
\delta p_{i}=[-\frac{\partial H}{\partial q^{i}}(q,p)-\lambda^{a}\frac{\partial \alpha_{a}}{\partial q^{i}}(q,p)]\delta t\\
\end{array}
\end{equation}
This would lead to one classical trajectory for $\lambda_{a}=0$  but for non-zero $\lambda$ we have two different trajectories described by different $\beta^{a}$. 
To generate quantum mechanics we take the extreme situation in which the solution $S(q^{i},t)$ does not contain any integration constant at all $(m=n)$. In that case, any two solutions contained in $S(q^{i},t)$ have the same values of a complete set of commuting conserved quantities, but differ for their conjugates. Basically an early form of position-momentum indetermination. 
In the case of constraint systems we can follow exactly the same approach as in the case of the theory of incomplete systems, which creates an interesting link between the implementation of constraints and the quantum theory, as can be seen in what follows. 
We identify the $\alpha_{a}$s with an abelian representation of the constraint surface. The conjugate variables $\beta^{a}$ are pure gauge, while $\alpha^{A}$ and their conjugates $\beta_{A}$ which commute with $\alpha_{a}$ form a complete set of gauge invariant functions. On this reduced phase space, these will be canonical coordinates. 
The generating function $S(q^{i},\alpha^{A},\alpha_{a},t)$ then defines a canonical transformation 
\begin{equation}
(q^{i},p_{i})\rightarrow \alpha^{A},\beta_{B},\alpha_{a},\beta^{a}
\end{equation}
and the constraints just become $\alpha_{a}=0$. Replacing this in $S(q^{i},\alpha^{A},\alpha_{a},t)$ one obtains a function $S(q^{i},\alpha^{A},t)$ which obeys the equations
\begin{equation}
\begin{array}{c}
G_{a}(q^{i}, \frac{\partial S}{\partial q^{i}})=0\\
\frac{\partial S}{\partial t}+H_{0}(q^{i},\frac{\partial S}{\partial q^{i}})=0\\
rank(\frac{\partial ^{2}S}{\partial \alpha^{A}\partial q^{i}})=n-m\\
\end{array}
\end{equation}

However, the information about $\frac{\partial S}{\partial \alpha_{a}}=-\beta^{a}$ is lost and the conjugate variable $\beta^{a}$ becomes arbitrary or undetermined. The equations above are the Hamilton Jacobi equations for a constrained system. 
If $(q^{i}(t), p_{i}(t))$ is at each time instant a solution of 
\begin{equation}
\begin{array}{c}
p_{i}=\frac{\partial S}{\partial q^{i}}\\
\\
\beta_{A}=-\frac{\partial S}{\partial \alpha^{A}}\\
\\
\end{array}
\end{equation}
one obtains
\begin{equation}
\begin{array}{c}
\dot{q}^{i}=[q^{i},H_{0}]+\lambda^{a}[q^{i},G_{a}]\\
\\
\dot{p}_{i}=[p_{i},H_{0}]+\lambda^{a}[p_{i},G_{a}]\\
\end{array}
\end{equation}
and the equation of the constraints 
\begin{equation}
G_{a}(q,p)=0
\end{equation}
Given a choice of the multipliers $\lambda_{a}$, we obtain a solution of the equations of motion $(q^{i}(t), p_{i}(t))$. Now, the Hamilton Jacobi function $S(q^{i},\alpha^{A},t)$ is gauge invariant as it contains all the solutions that are related by a gauge transformation to each other. 
A solution that realises the equations above including the constraints is called a complete solution of the Hamilton-Jacobi system. Such a complete solution describes gauge related trajectories that produce the same values for the complete set of gauge invariant variables $\alpha^{A}$ and $\beta_{A}$ and that differ only through the values of the pure gauge degrees of freedom $\beta^{a}$. The incomplete solutions here contain trajectories that are physically distinguishable which means they differ in the values of the gauge invariant quantities. In the extreme case, the solutions of $S(q^{i},t)$ that depend on no integration constant are specifically quantum. 
The unifying principle here is that for a solution of $S(q^{i},\alpha^{A},t)$ that depends on $n-m$ integration constants $\alpha^{A}$ it doesn't matter whether the fixed integration constants $\alpha_{a}$ are set to zero by our choice or due to a gauge-invariance principle. Also, it doesn't matter whether their conjugate momenta $\beta^{a}$ are unknown intermediate physical variables or are pure gauge variables. 
The main idea is that the indistinguishability of the conjugate momenta could appear due to a true gauge equivalence, or it could appear also because of a fundamental lack of knowledge, the first being the result of a gauge invariance, the second being the result of quantum mechanics. In both cases however, the system behaves the same. Therefore, and this completes this proof, gauge symmetry and quantum mechanics have a common origin. The only difference being in what we decide to consider a "fundamental indiscernibility" as opposed to a gauge symmetry between "truly" equivalent values. In any case this distinction is irrelevant. In the Feynman path integral approach, we are integrating across all possible paths that can occur in an experiment, with the property that, while we imagine them to be different, neither of them are directly measurable. The Feynman paths are continuous but not continuously differentiable, leading to the same indetermination we see in quantum mechanics. In the process, we have to bound together (re-classify) those paths that are "truly" equivalent in the sense of being gauge related. This is the principle of BV-BRST quantisation of gauge fields, and a whole technical mechanism for quantisation of such problems (open gauge algebra, etc.) has been devised. But from a physical point of view, the distinction between "truly unknown" distinctions and "gauge-undetermined" distinctions is meaningless. We cannot in principle measure neither, nor. So what is the distinction then, between gauge theory and quantum theory? Well, I can say that the distinction is that, in principle, if we add additional parameters and constraints we can make the one-to-many transformation invertible i.e. one-to-one. That is being said about gauge invariance. It is said that in principle we could eliminate it, by just adding extra parameters. Funny enough, the same is being said about quantum mechanics and its "hidden variables", although nobody successfully managed to actually do that. The same is true for gauge invariance. We are being told that in principle we could do without gauge degrees of freedom, but handling interactions in such a formalism is strange for the same reasons quantum mechanics with hidden variables is strange. Let's say we introduce a set of extra parameters that get rid of gauge invariance completely. What we obtain is a theory in which interactions appear as a mysterious action at a distance, in which the particle "knows" how to move in the presence of other particles in a globally defined way. If we know in advance what that globally defined way is, the problem is solved. The problem is, we usually don't. Exactly the same thing, but in a more acute way happens in quantum mechanics. If we put in hidden variables, the main problem is the non-localisation of interactions. Those variables seem to "know" in advance about choices of measurement devices and additional parameters introduced in the system that determine certain previously undetermined quantities, and can miraculously transmit them at a distance instantaneously. Of course, nothing of this sort actually happens, this remains just an artefact of the interpretation we are trying to give, namely that of extra variables that would describe the system. It therefore seems that we have to accept that in order to describe the evolution of a gauge or a quantum system, we require more parameters than the dynamical parameters or than whatever can explicitly be measured at any single point in spacetime. Those parameters may be fictitious but they are necessary. 
Of course, in general we accept quantum mechanics to be defined by a scale given by Planck's constant, but the uncertainty is not bound by this. Indeed, the non-determination of quantities can happen at any scale and it depends on the specific choice of some constants of motion as opposed to others, or, in more modern terms, the choice of a set of observables as opposed to others. 
Let's see how this can work in neural networks.
In a neural network we have a series of dynamics, learning, activation, input, as well as various re-classifications of those, all of them being described by means of equations of motion and hence by means of differential equations. How we choose the dynamical variables is a subject that remains at the latitude of the specific analysis. For example, it is possible to reduce the weight dynamics to a parametric optimisation according to a specific loss function. However, as shown in [19], the weight dynamics can be described in terms of variables of the differential equation, leading to a full differential equation determination of the dynamics of the neural network. If weights are being considered as variables, the differential equation determining the evolution of the neural network, including the learning phase, can be re-written in terms of differential equations, and in particular in terms of a Hamilton-Jacobi differential equation. The Hamilton Jacobi differential equation can have as solutions for the dynamics of a mechanical system a waveform, in particular the phase of the waveform is given by the action functional. At this level, the equation of motion is still not quantum, but we are getting closer. As explained earlier on, the complete solutions of the Hamilton Jacobi equation are not the only ones that present physical interest. We can construct solutions that depend only on fewer integration constants. Those incomplete integrals depend on what we assume about the conjugate constants of motion $\alpha^{i}$ and $\beta_{i}$. These are time independent canonical variables which in neural networks can include the input patterns, the neuron state, and the weight. Those are of course variables to begin with but they can be separated and represented by different timescales when considering the full dynamics. In this sense, they can be approximated as parameters and optimised in a variational sense (see first part of [19]) or can be represented as full dynamical variables (see second part of [19]). From the perspective of the Hamilton Jacobi equation however, if we can make the solutions depend on fewer integration constants we obtain the incomplete integrals, by setting the ones on which the solution doesn't depend to zero (for example). The incomplete integral $S(q^{i}, \alpha^{A},t)$ can therefore be written in terms of the complete integral $S(q^{i},\alpha^{A},\alpha_{a},t)$ as in
\begin{equation}
S(q^{i},\alpha^{A},t)=S(q^{i},\alpha^{A},\alpha_{a}=0,t)
\end{equation}
Once the $\alpha_{a}$ is set to zero the dependence of $S$ on $\alpha_{a}$ is lost and the conjugate variables $\beta^{a}=-\frac{\partial S}{\partial \alpha_{a}}$ is unknown and that means that the equations and the complete integral cannot determine a unique solution of the equation of motion. 

Let us consider the observables of a neural network, which are now functions of the weights and neuron states as variables
\begin{equation}
J(t)=J(t,y(t),W(t))
\end{equation}
The dynamics of an observable will be written as a surface $J=J(t, y, W)$ and the associated dynamical equation would be
\begin{equation}
D(t, y, W, J, \frac{\partial J}{\partial t}, \frac{\partial J}{\partial y}, \frac{\partial J}{\partial W})=0
\end{equation}
The Hamiltonian will contain the weights $W$ and their conjugate variables $M$, aside of $t$, $y$, and $\Delta$ the conjugate variable of $y$. 
\begin{equation}
\frac{\partial J}{\partial t}+H(t,y,\Delta, W, M)=0
\end{equation}
with 
\begin{equation}
\begin{array}{cc}
\Delta_{i}=\frac{\partial J}{\partial y_{i}}, & M_{ij}=\frac{\partial J}{\partial W_{ij}}\\
\end{array}
\end{equation}
This equation describes the evolution of a neural network, including the learning phase. If the solution $S(q^{i},t)$ abandons all dependence on integration constants then for any two solutions contained in $S(q^{i},t)$ that have the same values of a complete set of commuting conserved quantities, we will have different conjugates. In general this is the way in which arbitrary functions are being introduced in this dynamics, and they correspond to both gauge freedom and to quantum phase. From this point of view, indeed gauge dynamics and quantum dynamics involve the same type of arbitrary functions and are fundamentally similar phenomena. However, on the quantum side, the indetermination is forced on us by a limitation of defining certain simultaneous observables, while in gauge theory it is usually expected for the arbitrariness to be strictly a matter of choice. In reality, in both cases we have a fundamental limitation on the types of choices we can make. The exact same thing happens in the case of neural networks. However, here we may only set some integration constants and conjugate partners undetermined, leading to a dynamics of a constrained system with gauge. 
To see that something similar happens in the process of learning in a neural network, let us separate the dynamics into what has been learned in the past epoch, and what has to be learned in the present epoch
\begin{equation}
W_{ij}(nT+\tau)=W_{ij}((n-1)T+\tau)+\Delta W(nT+\tau)
\end{equation}
with $t=n\cdot T+\tau$, $0\leq\tau < T$, $n=0,1,2,...,\Delta W$. This denotes the variation of the weights during an epoch. The Hamiltonian associated to this evolution can be written as
\begin{equation}
H=\sum_{k}\Delta_{k}\cdot F_{k}(t,y,S_{T}W)+\frac{1}{2\omega}\sum_{k,l}\sum_{v=0}^{n+1}(S_{vT} M)^{2}_{kl}+E(t,y,W)
\end{equation}
with 
\begin{equation}
\begin{array}{c}
F_{k}=\frac{1}{\lambda}[-y_{k}+f_{k}(\sum_{j=-1}^{N} S_{T} W_{kj}\cdot y_{j})]\\
\\
(S_{vT}X)_{kl}(t)=X_{kl}(t-vT)\\
\end{array}
\end{equation}
where $X=W,M$. 
what we can do now is to artificially introduce a gauge invariance by providing a transformation to which the final action is not sensible. This transforms one of the dynamical variables into an undetermined dynamical variable, exactly as in the case in which we set it to zero back in the previous example. The associated conjugate will therefore become undetermined leading to an irrelevant constant of integration and therefore to a partial integral solution. 
Now, if we consider those variables that we eliminated in terms of dependence on the action, as constants of motion, we can re-write them as constant surfaces 
\begin{equation}
J(y,\alpha,t)=W(y,\alpha)-\mathcal{C}t=const(\alpha)
\end{equation}
and with this we obtain a wavefunction of the form 
\begin{equation}
\psi(y,t)=A(y,t)exp[-\frac{i}{\hbar} J(y,\alpha, t)]
\end{equation}
of course, following the precise analogy, we would identify a "Planck constant" which would be less fundamental but useful in defining the indetermination that emerges from our choice of gauge invariance. 
There is no reason why we cannot repeat this procedure of adding a gauge invariance for each integration constant, leading to an overall indeterminacy as is the case in quantum mechanics. 
At this point it is essential to note that with this state of the neural network, we will obtain the same type of entanglement that we obtain in quantum mechanics, and therefore neural networks could in principle be used to develop quantum computations, by making use of their dynamics alone. 
Also, as opposed to [35], we do not consider here any thermodynamical assumptions. This result does not emerge from a large and eventually undetermined number of neurons, nor from bringing any other quantity in the neural network to the thermodynamical limit. In fact, the quantumness of this approach relies only on the interpretation of the dynamics of the neural network as a solution to the Hamilton-Jacobi equation, as shown in [19] and on the generalisation of that approach to the case in which arbitrary functions are available and can be included in the dynamics. 

To be even more specific, let us have a look at how one solved the Hamilton Jacobi equation in general. 
Having the canonical equations by Hamilton
\begin{equation}
\begin{array}{cc}
\dot{q}=\frac{\partial H(z,t)}{\partial p},& \dot{p}=-\frac{\partial H(z,t)}{\partial q}
\end{array}
\end{equation}
with $z=(q,p)$ we want to find a transformation of coordinates $Z=(Q,P)=\Phi(z,t)=(\Phi_{1}(z,t),\Phi_{2}(z,t))$ having the same form for the equations of motion, but with a new Hamiltonian $K$ say
\begin{equation}
\begin{array}{cc}
\dot{Q}=\frac{\partial K(Z,t)}{\partial P},& \dot{P}=-\frac{\partial K(Z,t)}{\partial Q}\\
\end{array}
\end{equation}
Given that our neural network obeys Hamilton Jacobi equation and that its solution involves this type of solution based on canonical (symplectic) transformations that lead us to hamiltonians that do not depend on certain variables, making their conjugates constants of motion, also means that our neural network dynamics has an implicit symplectic structure which we can use further on. This will prove useful in a future article and is mentioned here just for the sake of consistency. 
If we can make $K$ independent of $Q$ then $P$ becomes a constant of motion and we can simply write the solution of the equations of motion as
\begin{equation}
\begin{array}{cc}
Q(t,Z_{0})=Q_{0}+\int_{0}^{t}\frac{\partial K(P_{0},\tau)}{\partial P}d\tau, & P(t,Z_{0})=P_{0}
\end{array}
\end{equation}
The original solution we can then obtain by the inverse transformation, $z=\Psi(Z,t)=\Phi^{-1}(Z,t)$. A canonical transformation is determined through the equation
\begin{widetext}
\begin{equation}
p(t)\dot{q}(t)-H(z(t),t)=-Q(t)\cdot \dot{P}(t)-K(Z(t),t)+\frac{d}{dt} F(q(t),P(t),t)
\end{equation}
\end{widetext}
The function $F$ is called the generator of the transformation. The equation is satisfied if
\begin{equation}
\begin{array}{c}
p(t)=\frac{\partial F(q(t),P(t),t)}{\partial q}\\
\\
Q(t)=\frac{\partial F(q(t),P(t),t)}{\partial P}\\
\\
K(Z(t),t)=H(z(t),t)+\frac{dF(q(t),P(t),t)}{dt}\\
\\
\end{array}
\end{equation}
and the canonical transformations become
\begin{equation}
\begin{array}{c}
p=\frac{\partial F(q,P,t)}{\partial q}\\
Q=\frac{\partial F(q,P,t)}{\partial P}\\
\end{array}
\end{equation}
If $F_{qP}=det(\frac{\partial F}{\partial q_{i}\partial P_{j}})\neq 0$ we can invert those equations solving for $P=\Phi_{2}(z,t)$ (at least locally). This also gives then $Q=\Phi_{1}(z,t)$. We get also $z=\Psi(Z,t)$ and solve $Q=F_{P}(q,P,t)$ for $q=\Psi_{1}(Z,t)$ and substitute in $p=\Psi_{2}(Z,t)$. Then the new Hamiltonian is defined by 

\begin{equation}
\begin{array}{c}
K(Z,t)=H(z,t)+\frac{dF(q,P,t)}{dt}=H(\Psi(Z,t),t)+\\
\\
+\frac{dF(\Psi_{1}(Z,t),P,t)}{dt}\\
\end{array}
\end{equation}

The equations of motion are invariant in form under this transformation. The invertibility condition $F_{qP}=det(\frac{\partial F}{\partial q_{i}\partial P_{j}})\neq 0$ however may not hold globally and then one may change the variables on which the generator may depend, in particular one may have to extend the space of variables to define a unique inverse function, as has been shown in the discussion on gauge theory. For example one may use a function $F_{1}(q,Q,t)$ that has $\frac{\partial^{2} F_{1}}{\partial q\partial Q}\neq 0$. The resulting equations would then be
\begin{equation}
\begin{array}{c}
p\dot{q}-H=P\dot{Q}-K+\frac{dF_{1}}{dt}\\
\\
p=\frac{\partial F_{1}}{\partial q}\\
\\
P=-\frac{\partial F_{1}}{\partial Q}\\
\\
H+\frac{\partial F_{1}}{\partial t}=K\\
\end{array}
\end{equation}
There is always a generator that can represent locally a given canonical transformation, where the variables it depends on are $q=(q_{1},...,q_{n})$ and another $n$ new variables $(P_{i_{1}},...,P_{i_{k}},Q_{j_{1}},...,Q_{j_{n-k}})$. If we require the generator to be smooth then the transformation is symplectic which makes the Jacobian of the transformation symplectic. If $n=1$ then the condition is $det(M)=1$. What we need is to determine the generator such that the resulting hamiltonian $K$ is independent of $Q$ such that the solution would be 
\begin{equation}
Q(t,Z_{0})=Q_{0}+\int_{0}^{t}\frac{\partial K(P_{0},\tau)}{\partial P}d\tau
\end{equation}
This leads us to 
\begin{equation}
H(q,\frac{\partial F(q,P,t)}{\partial q},t)+\frac{dF(q,P,t)}{dt}=K(P,t)
\end{equation}
This is also a Hamilton-Jacobi equation. The complete solution would be a solution of this equation given that it depends on $n$ parameters $P_{i}$ and has $det(\frac{\partial^{2} F}{\partial q\partial P})\neq 0$. An arbitrary function will extend the type of solutions granted that the above invertibility condition is not satisfied. 
Let us consider the case of separability of time dependence and hence introduce a hamiltonian as well as a generator $F$ that are not dependent on time. This results also in $K$ being time independent and we look for a similar solution. We express the problem in polar coordinates as 
\begin{equation}
\begin{array}{c}
(q,p)=(\phi,I)\\
(Q,P)=(\psi,J)\\
\end{array}
\end{equation}
with $\phi, \psi \in [0,2\pi]$ and $I, J\in[0,\infty)$. Let also $F(\phi, J)=\phi\cdot J+G(\phi,J)$ and we try to find $G$ from the Hamilton Jacobi equation 
\begin{equation}
H(\phi, J+\frac{\partial G(\phi,J)}{\partial \phi})=K(J)
\end{equation}
with the transformations defined by 
\begin{equation}
\begin{array}{c}
I=J+\frac{\partial G(\phi,J)}{\partial \phi}\\
\\
\psi=\phi+\frac{\partial G(\phi,J)}{\partial J}\\
\end{array}
\end{equation}
If $G$ satisfies the Hamilton-Jacobi equation for a function $K(J)$ then $J$ is a constant and $I=J+\frac{\partial G(\phi,J)}{\partial \phi}$ is an invariant torus in the phase space. 
Given the constructions of the Hamiltonian in the context of machine learning that I will do in the next chapters, it is useful to follow more of the theoretical constructions here and to see how to solve the problem for a perturbed Hamiltonian of the form 
\begin{equation}
H(\phi,I)=H_{0}(I)+\epsilon V(\phi, I)
\end{equation}
We take out the first term in the Taylor expansion of $H_{0}(J+G_{\phi})$ and write the Fourier series
\begin{equation}
G(\phi, J)=\sum_{m\in \mathbb{Z}^{n}}g_{m}(J)exp(im\cdot \phi)
\end{equation}
leading to 
\begin{equation}
\frac{\partial G(\phi, J)}{\partial \phi}=\sum_{m\in\mathbb{Z}^{n}} i m\cdot g_{m}(J)exp(i m \cdot \phi)
\end{equation}
Having 
\begin{widetext}
\begin{equation}
\frac{\partial H_{0}(I)}{\partial I}\cdot \frac{\partial G}{\partial \phi}=\epsilon V(\phi, J+G_{\phi})+[H_{0}(J+G_{\phi})-H_{0}(J)-\frac{\partial H_{0}(J)}{\partial J}\cdot \frac{\partial G}{\partial \phi}]+[H_{0}(J)-K(J)]
\end{equation}
\end{widetext}
and we obtain the coefficients by simply Fourier transforming the above quantity
\begin{widetext}
\begin{equation}
g_{m}(J)=\frac{i}{m\cdot \frac{\partial H_{0}(J)}{\partial J}}\frac{1}{(2\pi)^{n}}\int_{T^{n}}exp(-im\cdot \phi)[\epsilon V(\phi, J+\frac{\partial G}{\partial \phi})+H_{0}(J+\frac{\partial G}{\partial \phi})-H_{0}(J)-\frac{\partial H_{0}(J)}{\partial J}\cdot \frac{\partial G}{\partial \phi}]d\phi
\end{equation}
\end{widetext}
for $m\neq 0$ and the zero mode projection would be obtained by defining $K$ as the average 
\begin{equation}
K(J)=\frac{1}{(2\pi)^{n}}\int_{T^{n}}d\phi[H_{0}(J+\frac{\partial G}{\partial \phi})+\epsilon V(\phi, J+\frac{\partial G}{\partial \phi})]
\end{equation}
This type of solution allows us to define our wavefunction using a multivariate phase that has periodic type phase solutions and hence has various gauge (indeterminacy) directions that are however not linked to some "vanishing" procedure like taking a thermodynamic limit. 
Therefore the neural network retains from its Hamilton-Jacobi incomplete integral description a phase depending on a multivariate generating function that abandons full knowledge on a series of observables/dynamical variables. Due to this, the description of the neural network is made in terms of a wavefunction with a complex phase that generates potential correlations between non-observable potential outcomes, triggering the emergence of entanglement. 
In this sense the gauge degrees of freedom and the quantum phase appear as a unified concept, one being in some sense dual to the other. 
Given the wavefunction we obtain as 
\begin{equation}
\psi(y,t)=A(y,t)exp[-\frac{i}{\hbar}J(y, \alpha, t)]
\end{equation}
our incomplete generating function will not depend on a series of variables (the ones we set to zero) which are associated to conjugate momenta that remain fully undetermined. This will allow the wavefunction to play exactly the role we demand it to do in quantum mechanics, namely that 
\begin{equation}
|\psi(y,t)|^{2}=\rho(y,t)
\end{equation}
namely it will obey Born's rule and behave like a probability amplitude or a pre-probability. 
Therefore now we can expect for true entanglement to occur in the inner dynamics of a neural network. 
This has been achieved with local gauge freedom and small gauge transformations. The result was a standard quantum mechanical property of the neural network. In a future article it would be interesting to study the effect on neural networks of large gauge transformations (not linked to unity). What would be the quantum analogue of a large gauge transformation not connected to unity? And for that matter, what would be the neural analogue of such a gauge transformation? I hope to be able to answer those questions in a future article. 

\section{non-separability of a neural network}
The problem of describing the functioning of a neural network learning stage is stated usually as a Markov decision process (MDP) where we have a set of states $S$, a set of actions $A$, and the probability of a transition $P_{a}$ as well as a set of rewards given for executing an action $a_{t}$ given a state $s_{t}$. With no stochasticity, $P_{a}=1$ and we set up the goal for the agent to realise the policy $\pi(s_{t})=a_{t}$ such that we obtain a maximised reward 
\begin{equation}
E[\sum_{t=0}^{\infty}R(s_{t}, a_{t})|\pi]
\end{equation}
The policy is steadily learned, and we may introduce a discount parameter $\gamma$. The neural networks designed to learn the policy are constructed using parameterisations in the form weights and biases defined by $\theta$. The $Q$-value is given by 
\begin{equation}
Q(s_{t}, a_{t})=r_{t}+max_{a_{t+1}}Q(s_{t+1}, a_{t+1})
\end{equation}
and represents the numerical estimation of the reward after the agent performed the action $a_{t}$ at the state $s_{t}$ and $r_{t}$ is the reward at the step $t$ while $max_{a_{t+1}}Q(s_{t}, a_{t})$ is the maximum future value of the reward. Our neural network is the $Q$-function and the learning policy is defined for discrete actions spaces, being formulated as the policy parametrised by $\theta$ that would lead to the maximum $Q$ value. The Bellman equation is used to obtain the mean squared loss function and then to calculate the gradients needed for back-propagation
\begin{equation}
L_{t}(\theta)=E[(r_{t}+max_{a'}Q(s', a', \theta')-Q(s, a, \theta))^2]
\end{equation}
 Such an algorithm usually presents convergence problems because the maximisation can lead to over-estimations of $Q$. In order to correct that one may use separate target networks to predict the future $Q$ value inside the max operation, or dueling networks which have separate network heads that predict the advantage and value components of the Q-value, noisy nets, and others. 
In any case, the construction of the neural network, by means of a set of nodes linked by edges that provide inputs and outputs that are spread across a network have at least some properties which I want to underline here. 
First of all, the state produced by such a network is naturally non-separable. Hence it takes naturally into account one of the most fundamental aspects of entanglement, namely that the global system cannot be separated into its components without taking into account some form of common shared global information. 
Let us consider a neural network defined by an activation function $\sigma : \mathbb{R}\rightarrow \mathbb{R}$. This activation function may be a vector $\sigma(x)=(\sigma(x_{1}), \sigma(x_{2}), ... , \sigma(x_{3}))$. Let us consider a function $f:\mathbb{S}^{d-1}\times \mathbb{S}^{d-1}\rightarrow \mathbb{R}$ such that $f(x,x')=g(\Bracket{x,x'})$, for $g:[-1,1]\rightarrow R$. It has been shown [2] that the depth of the neural network gives a profound difference in whether the function can be approximated by a neural network. Indeed, $F:\mathbb{S}^{d-1}\times \mathbb{S}^{d-1}\rightarrow \mathbb{R}$ can be implemented by a depth-$2$ network of width $r$ and weights bounded by $B$ if 
\begin{equation}
F(x,x')=w_{2}^{T}\sigma(W_{1}x+W_{1}'x'+b_{1})+b_{2}
\end{equation}
with $W_{1},W_{1}'\in [-B,B]^{r\times d}$, $w_{2}\in[-B,B]^{r}$, $b_{1}\in[-B,B]^{r}$ and $b_{2}\in[-B,B]$. 
$F$ could be implemented by a depth-$3$ $\sigma$-network of width $r$ and weights bounded by $B$ if 
\begin{equation}
F(x, x')=w_{3}^{T}\sigma(w_{2}\sigma(W_{1}x+W_{1}'x'+b_{1})+b_{2})+b_{3}
\end{equation}
for $W_{1}$, $W_{1}'$ $\in [-B,B]^{r\times d}$, $W_{2}\in [-B,B]^{r\times r}$, $w_{3}\in[-B, B]^{r}$, $b_{1}, b_{2} \in [-B, B]^{r}$ and $b_{3}\in[-B,B]$. 
Polynomial-size depth-two neural networks with exponentially bounded weights will not be able to approximate $f$ whenever $g$ cannot be approximated by a low degree polynomial, however such functions can be approximated by polynomial size depth three networks with polynomial bounded weights. This gives a fundamental non-linear distinction between a depth $2$ and a depth $3$ neural network. This result however is not sufficient to prove non-separability. In order to do that, one has to also consider the agent-based learning process and the maximisation principle. This optimisation depends non-linearly on the possible cuts one could make in the network. If one reduces depth of a network one cuts through a tensorial product of the network components, leaving us with the results properly encoded by the distinction between a cartesian and a tensorial product. 
In fact it has been shown that a neural network with a single hidden layer with sigmoid activation functions and additional non-linearity in the output Neuron can learn ball indicator functions efficiently, while using a reduction technique, ball indicators cannot be approximated efficiently using depth-2 neural networks when the non-linearity in the output neuron is removed [3]. Therefore in some cases a larger neural network lacking some non-linearity in the output becomes less capable at learning functions than a smaller network with some output non-linearity. 
The depth of the network can be described as follows. Let us have a network in the form of a sequence of layers $\{L_{0}, L_{1}, ..., L_{N}\}$. The input is being provided at one side of the layers, and is transferred through the network which computes it as a sequential application of the layers
\begin{equation}
F(data)\leftarrow L_{N}(L_{N-1}...(L_{0}(data)))
\end{equation}
The loss function will be used to compute the gradients for the final layer $G_{loss}(output, label)$. The backpropagation over one layer will therefore be $L_{i}(\nabla)$ while the backpropagation over the entire network will be denoted $F^{T}(\nabla)$. Backpropagation over the entire network is represented by a sequential application of backward layers 
\begin{equation}
F^{T}(\nabla) \leftarrow L_{1}(L_{2}(T)... (L_{N}^{T}(\nabla)))
\end{equation}
The process of backpropagation allows the network to construct the suitable internal representations given by its sets of parameters, such that it can learn the mapping connecting the input to the output. The main part of the backpropagation method is the construction of the gradient of the loss function with respect to the weights of the network. The output of each neuron goes through an activation function, say $\sigma(x)=\frac{1}{1+e^{-x}}$ and combines all the previous results of the neural net, namely $\sum_{k=1}^{n}w_{kj}o_{k}$. As the derivative of the activation function is $\sigma'(x)=\sigma(x)(1-\sigma(x))$ the input of a neuron usually includes the weighted output of all the other previous neurons. 
If we think about this process, we will find some rather interesting connections to what we learn from quantum mechanics. Particularly, the way in which global information is being dealt with reminds us of quantum superposition of various forms. The only difference is the way we choose to interpret those outcomes. In fact, often the outcomes of a classifying neural network are probabilistic, in the sense that they provide probabilities for the correctness of the outcome of the classification problem. However, if we look at the internal mechanisms of the neural network, we notice a series of linear combinations followed by integration through the input branches of neurons that lead to superposed effects in the output. Each neuron comes with a non-linear component, namely the activation function, which allows the neural network to learn and take into account non-linear effects. What we obtain is a global characterisation of the state of each neuron in the form of a linear superposition, modulated by a non-linear activation function which introduces a threshold for the superposition to be considered. Needless to say, the signalling remains local in all situations. However, the resulting information integrates global data. While each neuron continues to access information via signals, locally, the repetition of the cycles results in global information being encoded in a non-separable way.
The backpropagation sends the information about the optimisation results back to the network, tracing back the layers and allowing for updates. This step is obviously crucial as, without it, no learning could take place. This inverse problem is actually one of the most investigated and optimised tool in machine learning. It is exceptionally important for the gradient descent methods to be well designed to make this process amenable to a rather large problem as the one generated by neural networks. However, another aspect of this method is to reinforce the non-separability of the network. Depending on the network depth, we have different learning capacities. The whole process of forward and back - propagation amounts to what we would call in quantum mechanics entanglement. Indeed, from this point of view, the classical neural network is a highly entangled state. How can that be even possible if the entanglement is basically a quantum property that relies on the non-realism principle described above? There could be no entanglement in systems showing only one single underlying fundamental outcome. However, we already went through the process of generating the neural network, and it appeared in several situations how this assumption stops being true. For any given input state, which finally is classical, as in, determined with one absolute value for its properties, the neural network does something interesting. It takes those input values, it generates a set of linear superposition between them, and it applies them to the next layer by means of activation functions. The activation functions cut some signals, while allowing others to pass, and the optimisation algorithm forces the network to gain access to the global structure of the problem manifold. Therefore, in the training phase, we are faced with a series of potential states of the system, each taken into account in each iteration of the loss function, and each being forced to take into account the global structure of our problem. This provides our network with a series of many potential outcomes, hence it lifts its state from one single value for its properties, to several possible values. Therefore, we could, in principle, describe the network as a whole by means of an operator-valued observable. We can also generate its potential eigenstates, by following closely the learning process. This still does not make the system truly quantum. What we need is non-separability. This is provided by the back-propagation of the gradient to make the optimisation/learning possible. At each step when the network learns, and therefore it submits its end-point parameters backwards through the network, it generates global information by means of its loss function, that could not be found at the previous step, when the signal did not yet advance up to the current layer. 

In terms of the Hamilton-Jacobi dynamics, we find ourselves in the situation in which we introduce by means of the inverse problem an arbitrary function for the case in which the inverse problem is not one-to-one, a situation that appears in the process of learning. The arbitrary function, be it regarded as a gauge freedom or as a quantum phase, is then propagated and allows for a broader type of optimisation, that wouldn't be possible if no such one-to-many relation and an arbitrary function was present. This evolution then allows the neural network to generate (and consider in the context of problem solving) additional intermediate states corresponding to the intermediate non-realised states in quantum mechanics, that in their turn allow for correlations that would not appear in any classical system. Therefore, neural networks will be able to solve, quite generally, problems inaccessible to a classical computer.

Of course there is the final step, which in quantum mechanics is the Born rule. In quantum mechanics we deal with complex wavefunctions representing expectation catalogues (expression I learned from Schrodinger, and I enjoy using it to describe what wavefunctions are). Those wavefunctions have complex phases that allow them to combine and form wave-patterns leading to the famous problems of quantum interference in double slits experiments, etc. To obtain the probability (density) out of those, we need to apply Born's rule, namely to calculate the absolute value or the norm of the final combination of complex wavefunctions. Neural networks also combine various branches of their inputs, leading to interesting interferences, but they do that by branching out "dendrites" for input signals and providing outputs. The outputs will really depend on how we decide to interpret those combinations. In most applications to strongly entangled quantum problems, those inner branches of a neural network are being interpreted as individual wavefunctions, and hence the process of combination inherent to a neural network, will construct proper entanglement which allows us to obtain accurate simulations of our quantum system. If however, we force the neural network to operate with classical probabilities, we will obtain classical results. Finally, the sole difference between a neural network and a quantum system is the way we decide to deal with the representation of data that we feed to the network and with the output. If what we feed to the network is a wavefunction, and we treat it as such, the neural network will optimise it in a quantum manner, giving a decisively better approximation to the quantum problem at hand. Of course, the process should not be mistaken for quantum computation. A quantum neural network proper, would do a far better job as it would harness the real power of entanglement for each of the branches, which would become quantum circuit lines. However, while not being a quantum computer, the neural network certainly has some remnant quantum properties that are inherent to its working style. 

 This remains valid in both biological and non-biological systems. The signals themselves always are transmitted in a local fashion, but the local signals end up containing more and more global information, leading to entanglement (i.e. non-separability). Finally, entanglement is also a causal and local phenomenon which however, encodes global information in a non-separable way. I must underline that quantum entanglement is a strictly causal and local phenomenon, its more-than-classical correlations being the result, among others, of having several possible outcomes for a property of the system, outcomes that are not realised unless a measurement capable of detecting them is defined and/or performed. There is no hidden variable or super-luminal transfer of information and because of that, we can say that entanglement remains a local phenomenon that encodes global information. Therefore in any case, both in biological and non-biological neural networks, the neurons will always access information locally, but that information may have a global component that results in the non-separability of the resulting informational content, a property shared by any neural network because of the optimisation/extremisation phase (also known as learning). 
 
Probably the most important aspect of this way of thinking is to understand what it means to be quantum. The non-separability of a space of states is essential, together with the linearity of the processes. The linear combinations are being performed by the neural network and integrated, but the process of interference is done via a non-linear function, the activation function. However, aside of that, the process has many similarities with a quantum system. Particularly, it generates entanglement, in the sense of generating non-separable state spaces on the network. This is being done by means of the back-propagation and learning sections of the method in the case of the simple neural network employed in the example below. This allows us to access the global information of the problem manifold. That information would not be retrievable in any separated piece of the network. This also explains some of the standing problems of neural networks, namely that it is usually hard or impossible to detect causal relations between the inputs and the outputs of the network, and that it is generally difficult to formulate local explanations of the methods leading to the outcome of the learning phase. This is so precisely because there is no local way of understanding the global problem, nor is there a strict causal connection, rather one that is based on weak-quantum correlations. This does not contradict the fact that neurons will continue to communicate among themselves in a strictly local and causal way. 

\par The fact that there exists a Hilbert space representation in the form of a Kernel for any learnable output of a deep neural network is known from [11]. There it is also explained that the parameters of the neural network can be regarded as superpositions of training examples in the state of the network, and hence the Kernel representation encodes all possible learned training examples as well. 
\par Let us look at the process described above by means of a simple system with one input neuron, one hidden neuron, and one output neuron. In the first phase we have the transition from our initial neuron to the hidden neuron, the information is being combined via a linear combination, and the non-linear activation function generates the output, and gives us access to its own parameters. The hidden neuron then carries the result of the activation function as an output towards the input of the next neuron or layer. There, another linear combination is performed, another non-linear activation function is introduced and the output is generated. If there was only this step of forward generation and propagation, then the neural network would be separable (at least in principle). Sure, reducing its size would affect the outcomes, but the global information would emerge only at the level of a one-step linear combination. However, we also have the learning phase. That particular phase generates the back-reaction where the gradient is being propagated in the direction of the blue wave-fronts in Figure 1. The parameters are being transferred by means of this gradient backwards via the blue arrows and a learning process is started. This phase introduces global information through the loss function that makes the neural network not separable, and makes the global state manifest, a state that cannot be recovered in the local neurons anymore. In quantum mechanics we call this entanglement. 
\begin{figure}
  \includegraphics[width=\linewidth]{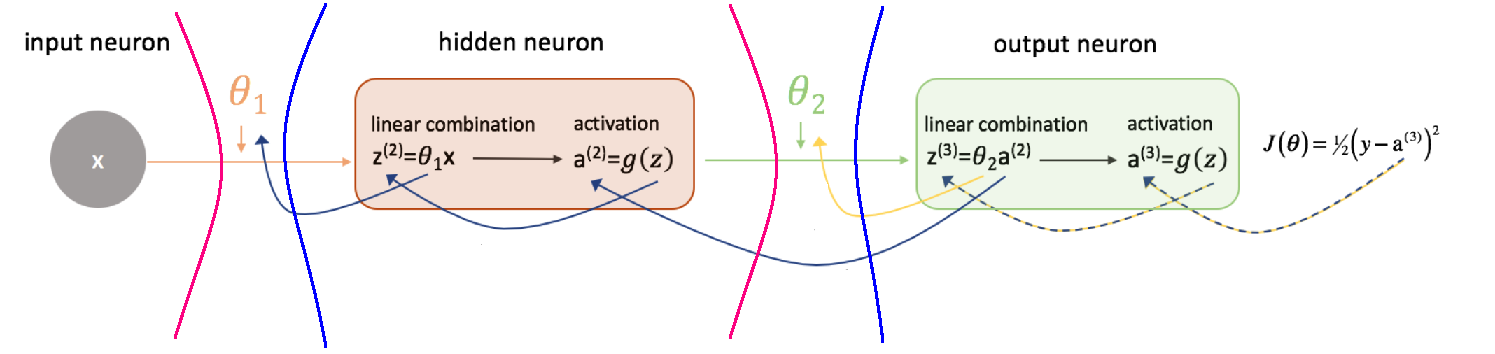}
  \caption{The evolution of information in a simple neural network. The red wavefront shows the direction in which information is separable. The gradient back-propagation however, drawn as the blue wavefront makes the global information on the network non-separable and the overall state space of the network more "quantum"}
  \label{fig:gauge1}
\end{figure}
The way in which this neural network is quantum is probably not the most obvious one, or one that is expected by physicists working with quantum mechanics at a different level. If one axiomatises quantum mechanics and generates a series of axioms that could define it completely as such, one may ask what would happen if some of those axioms are being abandoned? Quantum mechanics has its very special way of dealing with global information, namely it postulates, correctly so, that there exists global information that cannot be recovered locally. This is the foundation of entanglement. It also postulates the existence of catalogues of maximal knowledge (our wavefunctions) that can pre-interfere, before their results are being determined by properly constructed non-ambiguous observables. Both these aspects are being recovered in classical neural networks. The last one aspect that is not inherently built in the neural networks is the interpretation of the input and output information. In quantum mechanics we know the input information needs to be in the form of a complex expectation catalogue, while to obtain relevant output, we need to perform a Born-type procedure that will result in the correct probabilistic answer. It appears like quantum mechanics without this last axiom is found in neural networks.

\section{extracting the inner entanglement of a neural network, a practical approach}
Let us continue by asking a different question. If the inner state of a neural network is indeed an entangled, non-separable state, then it should be possible for a neural network to actually entangle input objects. The entanglement would then be extracted from the neural network into a pair of external particles. From a categorical point of view, the maps would therefore translate the entanglement rules into object entanglement. 
One of the first questions one may ask is how could we couple physical objects like photons or electrons with something usually considered just a computational tool, an abstract construction existing only as a computer software most of the time? In fact, even in a computer, one may ultimately think that the software creates a certain micro-circuitry representation of the functions being implemented, but in order for the process to be more practical, it would be useful to go back to reference [17] and also refer to ref. [33] and [34]. The main advantage is that a physically implemented neural network would be naturally behaving as such, therefore being practically much quicker than one simulated on a computing device. Moreover, such a physical neural network would indeed allow the input to be a physical quantum object, like a photon or an electron, leaving a second physical particle interacting with it, with the possibility to extract information as a neural network back-reaction, about the first interacting particle. If the inner workings of the neural networks are indeed lasers and mirrors, or microwave oscillators, performing the functions described above, the result should be a coupling between the two input particles (say photons) that, if the neural network had an entangled inner structure arising from its rules of functioning rather than from the quantum nature of the constituents, could be extracted as real measurable entanglement between the input particles. 

While the inner workings of a classical neural networks may resemble quantum entanglement or various other quantum information processing methods, one question emerges: is it possible to extract this quantumness from a neural network into the exterior? Otherwise stated, we may think at all the other interactions we know in high energy physics, for example electromagnetism, the strong interaction, the weak interaction, and even gravitation, and we can show that their mediators are indeed capable of entangling systems that they couple. Photons can certainly be used to entangle electrons via their mutual interactions, and decay products appear to be entangled due to the weak or strong mediating interactions between them. If a classical (from the point of view of the constituents) neural network could in principle entangle input objects then its quantum nature should be relatively obvious. This is indeed the case, at least at the level of the calculation I am presenting in this article. As opposed to the usual situation, however, entanglement would appear not due to an interaction mediated by a particle that is intrinsically quantum, but by a set of interactions in which the mediator is a classical neural network that would acquire quantum properties from its underlying rules of functioning.

First of all, let us consider a specific entanglement witness. In general an entanglement witness is a mathematical structure, usually a functional, that takes a given density matrix, and classifies the state it describes into entangled or separable [18]. 
For this we have to find a way in which we can describe the state of an out-going system, after interacting with a neural network, in terms of a density matrix. First of all, it would be interesting to understand the concept of interaction and what it would mean to interact with a neural network. An interaction is seen in general as an exchange of gauge fields in a given causal structure, bounded by the speed of light. From a fibre bundle point of view, a gauge field arises as a compensation to the fact that the connection across specific fibres has a globally defined curvature. In fact we can associate the field strength $F_{\mu\nu}$ of a gauge interaction with a curvature, and the associated transition function will require a change of the covariant derivative that will include our gauge field. All in all, a gauge field is therefore the response of nature to some inner space global structure, namely some inner curvature (in the case of strong, electromagnetic and weak interactions) or to some spacetime global structure (the curvature of spacetime in general relativity). That being said, in a neural network the process of backpropagation/learning is what is required to generate the global information that can then be used to construct the gradient of the loss function with respect to the weights of the network. This process plays a role similar to that of a connection but inside the neural network, and hence this is what we require in order to connect the two external systems we wish to entangle by this method. Basically this means that we have to transfer the backpropagation information from one side to the other and create an exchange (i.e. an interaction) with a system of two objects/particles that could be entangled. What are those objects? The easiest approach would be to consider another two neural networks, as they have the required internal structure that would make the interconnection the simplest. This could itself be an interesting result, but it wouldn't amount to much more than showing the internal quantum nature of a neural network. Rather than that, let me consider a way in which entanglement of two actual physical particles could occur via a neural network. For this to happen we have to somehow link the neural network and its backpropagation connection with the density matrix associated to the particles. Basically, the particles should be made to interact via the gauge connection defined by the backpropagation of the neural network. A physical neural network will employ real quantum objects, like photons, electrons, etc. In order to show that a classical neural network can entangle, care should be taken to keep those constituents in a non-entangled form. This can be done by employing the experimental devices proposed in [33], [34] as they can be tuned such that they continue to behave classically. To analyse the experiment we would require some tool that would encode the evolution of the density matrix via a neural network gauge interaction. 
It would be interesting to analyse the evolution by means of a master equation that could reveal the behaviour of entanglement as described from the point of view of an open system. 
We could therefore start with the quantum Boltzmann equation. The form usually taken is 
\begin{equation}
\frac{d\rho_{ij}}{dt}=i\Bracket{[H_{int}(0), \mathcal{D}_{ij}]}-\frac{1}{2}\int_{-\infty}^{\infty} dt\Bracket{[H_{int}(t), [H_{int}(0), \mathcal{D}_{ij}]]}
\end{equation}
where usually $H_{int}$ would be the interaction Hamiltonian while $\mathcal{D}_{ij}$ would be the number operator of the involved particles. For the number operator it is relatively easy to come up with an expression given that we talk about conventional particles, but for the interaction Hamiltonian the situation is somewhat more complicated as we need to deal with an expression relevant for the neural network. 
In general the Hamiltonian appears as a charge of the invariance of the system to time translations by means of Noether's theorem. We can consider the Hamiltonian action as
\begin{equation}
S_{H}[q,p]=\int_{t_{i}}^{t_{f}}[p\dot{q}-H(q,p,t)]
\end{equation}
take an infinitesimal shift in time
\begin{equation}
t'=t+\delta t
\end{equation}
with $q'(t')=q(t)+\delta q$, $p'(t')=p(t)+\delta p$ and requiring invariance under time translation after some very simple algebra we obtain $H(q,p,t)=const$ which defines our Hamiltonian function. In a neural network time steps are somehow different. In the problem of reinforced learning, there is an optimisation process in which agents interact with an environment and start developing a strategy that results in the optimum of some form of reward or cost function. The hamiltonian would play the role of a constant of "motion" in this context, resulting from the invariance of our "theory" to time translation. To construct something like this we have to pay attention to work at the level of the equivalent of the "action functional" where the extremisation would make sense, and to impose a time step variation in order to find the conserved quantity. Basically, an application of time-based Noether equation to neural networks. The result for all the actions taken would be an extremum over theta, hence
\begin{equation}
\frac{\delta \mathcal{S}_{t}}{\delta t}=0
\end{equation}
Even more interestingly, it has been shown in [19] that the activation and weight dynamics of neural networks behave in a way that can be described by means of a Hamilton Jacobi type partial differential equation, where the weights become either parameters or variables. If we take the weights as variables we can derive a Hamilton function that obeys a second order differential equation that encodes "forces" experienced by the 

weights in the presence of a potential encoded by learning laws or loss functions [20], [21]. 
But the learning laws are implemented in physical neural networks in the form of actual interactions and potentials, and hence carry the same structure, from the perspective of the in-coming and out-going particles (electrons, photons) as physical interactions. This brings us to the idea that our interaction Hamiltonian will be represented by some form of expanded loss function that can be represented in terms of the "gauge fields" described by the inner connection in the neural network.
The principle of this generalisation should be relatively clear from now on: we can write the inner dynamics of the neural network in terms of a Hamilton-Jacobi equation, which is particularly amenable to the construction of a quantum interpretation. This behaviour of weights and signals is then translated into entanglement for the particle physically interacting with the network. In reality, the Hamilton-Jacobi equation was a precursor to Schrodinger's equation with the sole difference that the wave-form results needed to be re-interpreted as probability amplitudes. Indeed, the Hamilton-Jacobi equations related for the first time trajectories to wavefronts, leading to the earliest forms of quantum mechanics. As we know, this duality was the same as the one in optics where the description of rays in geometric optics could be replaced (in some context) by the wavefront approach given by the Huygens principle. It is due to Hamilton that we now understand that this duality can be used to describe mechanical systems, where the collection of points reached by the light at time $t$, in optics (the "wave-front"), characterised by the traveling time, is replaced by our mechanical action. Now, in this article, we ought to understand that the same approach can also describe neural networks. 
In the case of a neural network, we also see the output of neurons or of layers of neurons as being a function of time, obeying a certain dynamics. This dynamics is determined by the network topology, the input patterns, and the weights dynamics of the system. Given the neuron or neuron layer output, $y(t)$, the input $in(t)$ and the weights $W(t)$ we can write 
\begin{equation}
\frac{d}{dt}y(t)=F(in(t), y(t), W(t))
\end{equation}
and the weights dynamics is described by 
\begin{equation}
\frac{dW}{dt}(t)=G(in(t), y(t), W(t))
\end{equation}
The time scales of the two dynamics can vary, with the activation dynamics usually being slower than the weight dynamics. However, as shown in [19] a separation of time-scales is not necessary and the activation, weight, and learning dynamics can be studied in a unified manner. The learning process as a whole becomes a dynamical process obeying a Hamilton Jacobi equation, with a Hamiltonian function and an equivalent wave dynamics reminding us of quantum mechanics. The Hamiltonian derived following [19] will contain the information about the network's topology and the learning method chosen. The learning method will also define the connection on the neural network leading to the evolution of the neural network "gauge field" dynamics and its ability to entangle. We could go to the next step and define a truly quantum formulation of the hamiltonian and obtain a quantum Boltzmann equation which then can be solved to follow the entanglement dynamics. The solution of the initial network differential equation is interpreted as an accumulated error surface over the state space. If weights are introduced as variables in the dynamical system describing the neural network, the corresponding partial differential equation is a Hamilton Jacobi equation. The dynamics of the weights is completely determined by the Hamiltonian. 
Inputs to the network can be either reference signals for learning, actual input signals, as well as weights. Those form a boundary condition to the neural network through which it interacts with the exterior. The action of the network produces a set of observables $J(t)$ which integrates the states of the network and the boundary conditions. This leads to an internal superposition which enables entanglement. We can define them as $J(t)=J(t,y(t))$. The mean error is an example of such an observable for supervised learning. Those observables form a trajectory in the state space. Keeping the weights fixed and varying the rest of the boundary conditions leads to a new trajectory and so on, leading to a surface we call $J=J(t,y)$ which encodes the full information about the dynamics of the network with a given set of weights. To describe the structure of the surface we will also employ the derivatives of $J$ with respect to $t$ and $y$ and hence the surface will obey an equation of the form 
\begin{equation}
D(t, y, J, \frac{\partial J}{\partial t}, \frac{\partial J}{\partial y}, \frac{\partial^{2}J}{\partial t^{2}}, \frac{\partial^{2}J}{\partial y^{2}},...)=0
\end{equation}
Simplifying the problem leads to a restriction of the $D$ operator yielding 
\begin{equation}
\frac{\partial J}{\partial t}+h(t,y,\Delta)=0
\end{equation}
where $\Delta_{i}=\frac{\partial J}{\partial y_{i}}$ and $\Delta$ plays the role of a time varying form of the back-propagation hence becoming particularly important for our entangling process. 
This is a Hamilton Jacobi equation which in optics links the wavefront to the trajectory and which lies at the foundation of quantum mechanics given a re-interpretation of the wavefront expressions as underlying probability amplitudes. This is just another way of saying that we have a consistent wavefunction interpretation of the neural network. It is worth mentioning that by itself a Hamilton-Jacobi equation is not a Schrodinger equation as used in quantum mechanics. In fact we can simplify the Schrodinger equation to obtain a Hamilton Jacobi equation, but we do need to add the fundamental Planck constant and re-interpret the results as wavefunctions to obtain Schrodinger's equation. The difference between the two is closer of being an interpretational one, except for the Planck constant which is fundamental to Schrodinger's equation. However, non-separability as obtained in quantum entanglement can be obtained from an overall adoption of the neural network and its inner workings as a starting point. In that sense, if we feed the neural network an input in the form of wavefunctions, it will be able to entangle them. Therefore what we "feed" our neural network must be quantum objects, that could in principle entangle, although their initial state should be separable. 
The Hamilton Jacobi equations give rise to a fundamental set of ordinary differential equations, also known as characteristic equations
\begin{equation}
\begin{array}{ccc}
\frac{dy_{i}}{dt}=\frac{\partial h}{\partial \Delta_{i}}, \frac{d\Delta}{dt}=-\frac{\partial h}{\partial y_{i}}, \frac{dJ}{dt}=\frac{\partial J}{\partial t}+\sum_{j}\Delta_{j}\cdot \frac{\partial h}{\partial \Delta_{j}}
\end{array}
\end{equation}
The process of solving these involves finding the solutions for $y$ and $\Delta$ and then integrate the third equation for $J$. 
If the weights are allowed to change as well, we obtain a set of fluctuating surfaces.
We can now define the Hamiltonian which is at most linear in the conjugate variable $\Delta$ and because $\frac{dy}{dt}=\frac{\delta h}{\delta\Delta}$ we can introduce a Hamiltonian for a neuron model
\begin{equation}
h=\sum_{j}\Delta_{j}\cdot F_{j}(t,y;W)+E(t,y;W)
\end{equation}
where $F$ represents the model, and we can already remember it again here
\begin{equation}
F_{i}=\frac{1}{\lambda}\cdot [-y_{i}+f_{i}(\sum_{j=-1}^{N}W_{ij}\cdot y_{j})]
\end{equation}
with the notation $W_{i-1}\cdot y_{-1}=Y_{i}$ the individual external input of a neuron and $W_{i0}\cdot y_{0}$ the neuron individual threshold. 
$f_{i}$ is the transfer function and $N$ the number of neurons. $E$ is an error function or a generalised cost or a Ljapunov function. $f$ also reflects the topology of the network and combined to $1/\lambda$ we obtain the time constant of a neuron. 
We obtain 
\begin{widetext}
\begin{equation}
\begin{array}{cc}
\frac{dy_{i}}{dt}=F_{i}(t), &\frac{d\Delta_{i}}{dt}(t)=\frac{1}{\lambda}[\Delta_{i}(t)-\sum_{j}\Delta_{j}(t)\cdot \dot{f}_{j}(t)\cdot W_{ji}(t)]-\frac{\partial E}{\partial y_{i}}(t)
\end{array}
\end{equation}
\end{widetext}
and 
\begin{equation}
\frac{dJ}{dt}=-h+\sum_{j}\Delta_{j}\cdot \frac{\partial h}{\partial\Delta_{j}}=-E 
\end{equation}
leading to 
\begin{equation}
J(t)=J(t_{0})-\int_{t_{0}}^{t}E(\tau, y(\tau); W(\tau))d\tau
\end{equation}
The next step would be to extend the Hamiltonian and the associated dynamics with respect to the weights and hence to replace the optimisation procedure as well with a causal dynamics. Weights are therefore shifted from being parts of the boundary conditions to being variables and the observables therefore become $J(t)=J(t, y(t), W(t))$ and then the dynamics of an observable is represented by a single surface depending on the network and the weights, leading to a fundamental equation of the form 
\begin{equation}
D(t, y, W, J, \frac{\partial J}{\partial t}, \frac{\partial J}{\partial y}, \frac{\partial J}{\partial W})=0
\end{equation}
We will have to introduce the weights $W$ together with their conjugated variables $M$, and we obtain the Hamilton Jacobi equations
\begin{equation}
\frac{\partial J}{\partial t}+H(t,y,\Delta, W, M)=0
\end{equation}
given 
\begin{equation}
\begin{array}{cc}
\Delta_{i}=\frac{\partial J}{\partial y_{i}}, & M_{ij}=\frac{\partial J}{\partial W_{ij}}
\end{array}
\end{equation}
and we extend the set of characteristic equations associated to the Hamilton Jacobi equation 
\begin{equation}
\begin{array}{cccc}
\frac{dy_{i}}{dt}=\frac{\partial H}{\partial \Delta_{i}}, & \frac{\Delta_{i}}{dt}=-\frac{\partial H}{\partial y_{i}}, &\frac{dW_{ij}}{dt}=\frac{\partial H}{\partial M_{ij}}, & \frac{dM_{ij}}{dt}=-\frac{\partial H}{\partial W_{ij}}\\
\end{array}
\end{equation}
and
\begin{equation}
\frac{dJ}{dt}=\frac{\partial J}{\partial t}+\sum_{j}\Delta_{j}(t)\cdot \frac{\partial H}{\partial \Delta_{j}}(t)+\sum_{ij}M_{ij}(t)\cdot \frac{\partial H}{\partial M_{ij}}(t)
\end{equation}
and here the surface $J(t,y,W)$ is generated by the manifold of all solutions obtained by varying the initial states of the system and the boundary conditions. 
It is not surprising that those parts of physics have elements in common: the duality between wavefront and trajectory in optics links information from all the domains reached by a wavefront and binds them to the overall trajectory of the ray leading to an optimised path associated with the classical extremum, while in quantum mechanics fluctuations around that path as well as non-perturbative contributions play a significant role. Neural networks in the weight space do a similar thing, they explore possibilities in the process of optimisation, leading to a classical optimum and fluctuations from around it. Therefore they integrate information from the weight space in a broader sense, and are amenable to a description by means of the Hamilton-Jacobi equation. It is totally not surprising therefore that a link between classical neural networks and quantum information can be established. 
The Hamiltonian obtained for the neural network is also generating a learning law hence the whole process through which the neural network learns is a result of the dynamics governed by the Hamiltonian and ultimately by our Hamilton Jacobi equations. Integrating we obtain 
\begin{widetext}
\begin{equation}
J(t)=J(t_{i})+\int_{t_{i}}^{t}(\sum_{j}\Delta_{j}(\tau)\cdot \frac{dy_{j}}{d\tau}+\sum_{ij}M_{ij}(\tau)\cdot \frac{dW_{ij}}{d\tau}-H(\tau))d\tau
\end{equation}
\end{widetext}
The variation of $J$ equated to zero leads to $\Delta_{i}(t_{f})=0$ and $M_{ij}(t_{f})=0$, and hence a choice of Hamiltonian leads to an extremal trajectory meaning that the Hamiltonian dynamics makes the neural network learn. 
We can decompose the weight functions into what has been learned in the past and what needs to be learned in the present 
\begin{equation}
W_{ij}(nT+\tau)=W_{ij}((n-1)T+\tau)+\Delta W(nT+\tau)
\end{equation}
with $\Delta W$ being the change of the weights in the current epoch. The associated hamiltonian becomes
\begin{widetext} 
\begin{equation}
H=\sum_{k}\Delta_{k}\cdot F_{k}(t, y, S_{T}W)+\frac{1}{2\omega}\sum_{k,l}\sum_{\nu=0}^{n+1}(S_{vT}M)^{2}_{kl}+E(t,y,W)
\end{equation}
\end{widetext}
where we can expand the expressions for $F$ and $S$ as in 
\begin{equation}
\begin{array}{c}
F_{k}=\frac{1}{\lambda}[-y_{k}+f_{k}(\sum_{j=-1}^{N}S_{T}W_{kj}\cdot y_{j})],\\
\\
 (S_{vT}X)_{kl}(t)=X_{kl}(t-vT), X=\{W, M\}\\
\end{array}
\end{equation}
Due to the analogy with mechanics, the first and third term in the Hamiltonian are associated to a potential called the "learning potential". This defines essentially our interaction term
\begin{equation}
H_{int}=\sum_{k}\Delta_{k}\cdot F_{k}(t,y, S_{T}W)+E(t,y,W)
\end{equation}
The equation of motion derived from this Hamiltonian is called a "learning law" in analogy to "Netwon's laws". It is particularly important to note that the operation $S_{T}$ has the role of returning to a previous temporal epoch, in the sense that 
\begin{equation}
(S_{\nu T}X)_{kl}=X_{kl}(t-\nu T)
\end{equation}
not only for the case of the overall optimisation function, but also for the potential at hand, leading to the 
creation of a time-wise correlated structure and potentially to a time-wise entangled system [22]. 
Usually, one would say, now is the right moment to introduce some quantisation prescription and to expand some generalised momenta and positions to the level of quantum non-commuting observables. That is however not the main goal of this work. I do not want to quantise a neural network, I want to show that it already has enough quantum remnants, classical as it is, to induce entanglement. In reality, transitioning from the Hamilton-Jacobi equation to Schrodinger equation requires a few changes in interpretation that are not included in the direct derivation of the Hamilton Jacobi equation, but appear only at the level of the types of solutions searched. The interpretational difference is that instead of dealing with the solutions of the classical Hamilton-Jacobi equation, we deal with the wavefunction as a probability amplitude and the action functional of Hamilton Jacobi becomes the phase of our quantum solution. Again, the neural network gives us both options. Depending on what we insert as an input, we may obtain classical probabilities, or quantum probability amplitudes. Sure, in principle all the dynamics can be derived equivalently via the Hamilton mechanisms, the Lagrangian mechanisms, or the Hamilton Jacobi mechanisms, however, the fact that neural networks are intuitively closer to the Hamilton Jacobi mechanism brings it closer to what we want to obtain as "quantum remnants" in neural networks. 
Now, $\Delta$ is the analogue of our momentum, as it results from $\Delta_{i}=\frac{\partial J}{\partial y_{i}}$. 
Therefore by the prescription of the minimal coupling it will contain the gauge field, and hence $\Delta$ will be deformed to include the gauge connection over our network $\Delta_{i}\rightarrow \Delta_{i}+c\cdot \dot{y}_{i}\cdot A_{i}$ with $A_{i}$ the associated gauge field and $c$ is some form of charge. However the "coupling" itself between the exterior fields (say photons) and the neural network is basically represented by feeding the neural network the quantum state of the initial particles in the form of a density matrix. It is interesting to note that we can indeed define a gauge connection on the neural network and hence an effective gauge field. After solving numerically for a one layer and two layer simplified case and taking an average, the potential produced by the neural network was similar to the Lenard-Jones potential, however in time evolution. That can be interpreted as the evolution in time of the learning phase of the network, which will have an extremum (minimum) time region where the information fed to the network by the first particle will co-exist in the network during the learning phase with the second particle. The obtained potential is represented in figure 1. It is very likely that the specific topology of the network will change this very simplified shape, but in any case, it shows that it is possible to get a sufficiently bound "interior state of knowledge" in the network capable of creating some correlation and, by the construction of the feedback mechanism, to actually correlate that information with the information about the real system that has been fed to the network. The exterior particle will receive a back-reaction from the network capable of entangling them. Moreover the potential shows us a competition between early time(or in this case negative time, as the time of early training, negative due to my own choice of the initial time convention) repulsion, and intermediary time attraction as well as a stable bound state at some extremum encoding the region where global information of the two particles becomes inseparable and hence entangled. 
\begin{figure}
\centering
\includegraphics[scale=0.5]{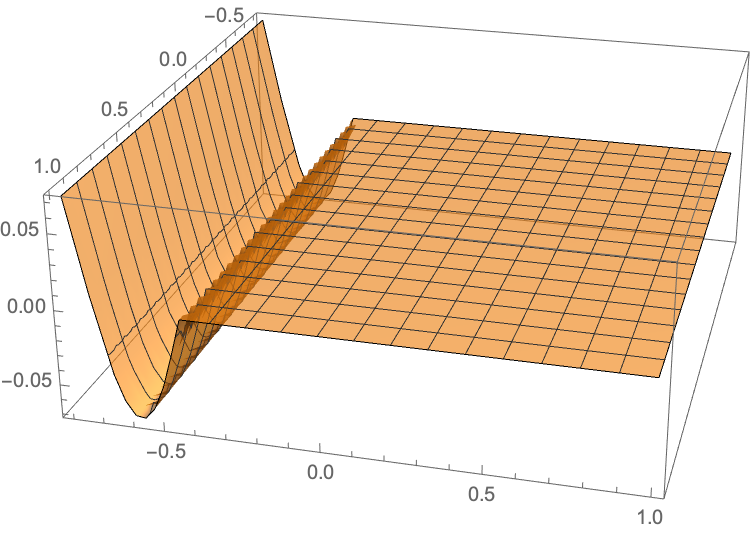}
\caption{The averaged interaction potential for a two layers network as evolving in time. The negative values for time are due to the conventional choice of the initial moment of input}
\end{figure}
Now let us see what happens to the system itself, in contact with a neural network. From the form of the potential for various layers, and due to the $S_{\nu T}$ operation we notice that the system is not separable. It is in a sense expected that the history of the neural network will play a role, as it is through it that it will learn. However, from the perspective of the Hamiltonian description this leads to what we would describe as a correlation sensitive channel. 
So, we now have an interaction Hamiltonian that is fundamentally not quantum, and I am not going to impose any quantisation rule on it. That will make the situation rather dull if the structure of the neural network was indeed purely classical. This however is not the case. While I do not impose from outside any quantisation rule, it is clear that the terms $\Delta_{k}(s,t)$ as well as $F_{k}(s,t)$ and $E(s,t,W)$ are operator valued. How come? Simply due to the nature of the neural network and the rules by which it operates. The index $k$ is counting the neurons while each object is then interlinked in layers with the rest of the system giving us an operator valuation of the respective objects. This means that we can still construct an operator representation of our momentum-analogue, position analogue, and Hamiltonian analogue objects and start linking them into an interaction picture. It remains to be seen whether they obey some analogue of the commutation relations. In fact, given the Hamilton Jacobi type of equation we are dealing with, we easily obtain the Poisson bracket relation between momentum and position in this context as well, with the observation that a translation towards the quantum mechanical commutation rules emerges simply if we decide to consider the objects described by the network as being quantum amplitudes. Therefore, without imposing any quantisation rule, but just deciding to consider the information as quantum, and hence without making any physical changes to the classical network, we can consider it, at least for this problem, a classical neural network with relevant quantum remnants. This simplifies the problem substantially because that means we just have to deal with a simple interaction problem, in a rather special potential that looks like the Lennard Jones potential, only not spreading spatially, but instead temporally. Be as it might, it is quite suggestive to the process to be described: two particles evolve, are being considered as input for a neural network, hence their wavefunctions superpose within the neural network, the neural network acts as the temporal potential described above, while the particles separate and continue their movement after being affected by the neural network back-reaction. The claim is that the particles just by being "observed" by the neural network (hence feeling the back-reaction of the network), become entangled themselves. 
This runs against a relatively fringe point of view on quantum mechanics where the idea is promoted that an "observer" or some "consciousness" (hence a neural network) would "collapse" the wave-function leading to a non-unitary process and the emergence of one certain outcome. While I am not a supporter of this viewpoint, it comes a bit as a surprise to notice the precise opposite: a classical neural network (or, in a more flamboyant language, and maybe imprecise sense, a "consciousness") would not only not "collapse" any quantum information, but it would even strengthen the correlation of the objects it is fed into as input. I find this observation intriguing. 
But let's leave those metaphysical musings aside and let's consider the actual behaviour of our system. 
For this we will employ the quantum Boltzmann equation 
\begin{widetext}
\begin{equation}
(2\pi)^{3}\delta^{3}(0)(2k^{0})\frac{d\rho_{ij}(k)}{dt}=i\Bracket{[H_{int}(0), \mathcal{D}_{ij}]}-\frac{1}{2}\int_{-\infty}^{\infty}dt\Bracket{[H_{int}(t),[H_{int}(0),\mathcal{D}_{ij}(k)]]}
\end{equation}
\end{widetext}
where the density matrix encodes the properties of the particles that are interacting that may be subject to entanglement, for example it could be a polarisation density matrix for photons where the elements are the respective Stokes parameters. 
Here we have to make a clear distinction between the interaction Hamiltonian which refers to the neural network, as it mediates the interaction with the outlying photons and the particle number operator which refers to the interacting particles
\begin{equation}
\mathcal{D}_{ij}(k)=a_{i}^{\dagger}(k)a_{j}(k)
\end{equation}
with the usual definition for the creator and annihilator operators. In general now, an operator $O$ will have an expectation value given by 
\begin{equation}
\Bracket{O}=Tr[\rho\cdot O(k)]=\int\frac{d^{3}p}{(2\pi)^{3}}\Bracket{p|\rho\cdot O(k)|p}
\end{equation}
with $\rho$ being the density operator
\begin{equation}
\rho=\int\frac{d^{3}p}{(2\pi)^{3}}\rho_{ij}(p)\mathcal{D}_{ij}(p)
\end{equation}
We have two terms, a forward scattering term and a damping term. Clearly the forward scattering term would be dominant, the others appearing in decreasing order due to the perturbative approach to the derivation of this equation. However, those terms are important in the process of entanglement. Traditionally such Boltzmann equations describe open thermodynamic systems that are not in equilibrium, and also traditionally, they do not properly take into account quantum effects. This is because traditionally one doesn't use the quantum density matrix as an operator propagated by the equation. In this context however, the evolution of the quantum density matrix will be able to detect entanglement dynamics both through the neural network and through any potential thermal bath the system may be in contact with. I am not considering that type of evolution here, as it would complicate things, but I expect that the entanglement effect will be diminished by it, without being fully eliminated. Once the input describing the state of the incoming particles is fed to the neural network, there will be an interaction between the neural network and the in-coming particles. Their properties will become correlated due to the neural network. But will they be entangled? Now that we have a dynamical density matrix, we can determine this by means of an entanglement witness. The calculations show clearly that there will be some superposition, but will there be entanglement? There are several entanglement witnesses that can determine this for us. Surely to find a Hermitian operator capable of detecting entanglement is possible, as some quantum information theorem assures us, however, it is an NP-hard problem to solve the separability problem, and hence we cannot really expect to develop an efficient and systematic classical algorithm capable of doing this. I will focus on a relatively famous entanglement witness called "concurrence". Its definition is 
\begin{equation}
C(\rho)=\sqrt{2(1-Tr[\rho^{(A)^{2}}])}
\end{equation}
This looks relatively complicated but if the system can be reduced to a $2\times 2$ case when concurrence has a simple form 
\begin{equation}
C(\rho)=max(0, \lambda_{1}-\lambda_{2}-\lambda_{3}-\lambda_{4})
\end{equation}
where $\{\lambda_{i}\}_{i=1}^{4}$ are the eigenvalues of $M=\sqrt{\sqrt{\rho}\tilde{\rho}\sqrt{\rho}}$ where
$\tilde{\rho}=\sigma_{y}\otimes \sigma_{y}\rho^{*}\sigma_{y}\otimes \sigma_{y}$ and the eigenvalues are in descending order.
An alternative would be the negativity 
\begin{equation}
\mathcal{N}(\rho)=\frac{||\rho^{T_{B}}||_{1}-1}{2}
\end{equation}
and many others. 
Now, the potential as a gauge field is in $\Delta_{k}(s,t)$ while the input which encodes the external particles is in $F_{k}(s,t)$. But in there, the input is being spread across the neural network by means of a series of linear superpositions at the level of each layer, and combined with global information by means of backpropagation. It is however combined also with the input from the other particle which is also backpropagated and becomes globally combined. This type of backpropagation I showed to be non-separable. 
Therefore the first term we need to consider will be
\begin{equation}
[H_{int}(0), \mathcal{D}_{ij}]=[\sum_{k}\Delta_{k}\cdot F_{k}(t, y, S_{T}\cdot W)+E(t,y,W), \mathcal{D}_{ij}]
\end{equation}
In a first approximation the non-commutative nature of the neural network will be manifest in the backpropagation phase, and that is encoded by $\Delta_{k}$. The rest of the terms are assumed to pass through the commutation rule. Considering the passage through the neural network as a global function, we can write
\begin{equation}
[\Delta + E(t,y,W),\mathcal{D}_{ij}]=[\Delta,\mathcal{D}_{ij}]
\end{equation}
for which we will need a symbolic evaluation
\begin{equation}
\Bracket{[H_{int}(0),\mathcal{D}_{ij}]}=Tr(\rho\cdot[H_{int}(0),\mathcal{D}_{ij}])
\end{equation}
and with 
\begin{equation}
\rho=\int\frac{d^{3}k}{(2\pi)^{3}}\rho_{ij}(k)\cdot \mathcal{D}_{ij}(k)
\end{equation}
leading to 
\begin{equation}
Tr(\int\frac{d^{3}k}{(2\pi)^{3}}\rho_{ij}(k)\mathcal{D}_{ij}(k)[\Delta(0),\mathcal{D}_{ij}])
\end{equation}
The next term is significantly smaller but connects the interaction term at the initial time and at a general time, followed by an integration over all time steps
\begin{equation}
-\frac{1}{2}\int_{-\infty}^{\infty}dt\Bracket{[H_{int}(t),[H_{int}(0),\mathcal{D}_{ij}]]}
\end{equation}
The inner commutator is known to be dependent on the overall backpropagation commutator $[\Delta, \mathcal{D}_{ij}]$ but the next commutator will be between this term and the future evolution of $\Delta(t)$, the result being then integrated over all times. This dependence on epochs has been implemented via $S_{T}$ and the evolution of $\Delta$ in time obeys the differential equation mentioned previously. The non-commutativity of the propagation of the information of the incoming particles and the back-propagation of the signal is assumed as a normal result of the construction of the network. As is known, by propagating forward one follows a chain of activating neurons, while the backwards process contains global information about the weights optimised in the process. It shouldn't appear as a surprise that the result will be non-commutative, and in fact it will result in a growing concurrence. 
This equation by itself is quite complicated, therefore I simplified it as much as possible at the level of implementation. In particular I only used one layer of weights, as a linear combination in the resulting equation. I also coupled the variables over adjacent time-steps as resulting from the estimations of the commutation rules. I then obtained a set of density matrices at various time-steps, I calculated for each of them the concurrence and plotted the result, as shown in figure 2. I cut the low time behaviour due to the unreliable nature of the early steps of the learning potential. Also, more testing with more suitable model equations could reveal details that are not visible here. Further work on more realistic neural networks (aside of the simple manually implemented linear combination of variables used here) may reveal new interesting patterns even at early times. I hope to present that in a future article. There have been many simplifications: the commutation rules are assumed in the $\Delta$ function as a result of the non-commutativity of the information flowing from the input end to the output end of the network and back. The coupling with another particle is assumed to operate in the same way, at the input of the neural network, hence the interaction of the two particles should be somehow separated in time. In any case, it appears that the coupling over different times does indeed generate entanglement. 
\begin{figure}
\centering
\includegraphics[scale=0.5]{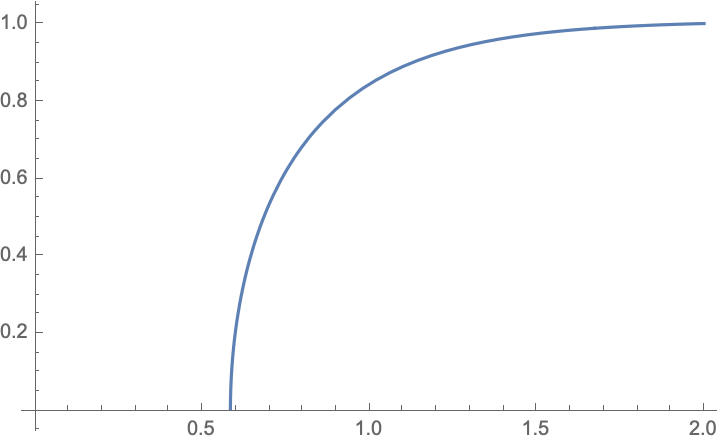}
\caption{The evolution of concurrence over a series of time iterations of the density matrix}
\end{figure}
\section{conclusion}
In conclusion what we have seen is that a classical neural network can actually entangle objects that it receives as inputs. This should not be surprising as the way a neural network works is to create a series of superpositions of potential states of the input system and then through backpropagation to create a non-separable structure which, when it interacts with the other particle, will encode in it global information about all possible states of the input (first particle). Backpropagation as weight optimisation is essential, although the weight optimisation can also be achieved by other means, for example via some annealing procedure. 
Involuntarily, these results also go against the philosophical belief that "consciousness collapses the 
wavefunction" [23], [24], [25]. A neural network (which is some very rough approximation for the, as of now, imprecise term of "consciousness") doesn't "collapse" any wavefunction, instead it has the possibility of further entangling objects it gets as inputs. The problem is of course poorly stated in this context. Of course any observer will observe only one outcome (you don't see a dead cat that is also alive), it is just by repeating the observations that the statistics will emerge which will obey the results of quantum mechanics, including state interference and entanglement. If that weren't so, we wouldn't be able to witness entanglement to begin with. But the goal of this article is not to solve poorly stated philosophical questions. 
It is interesting to note that neural networks can entangle physical systems. Most neural networks we use are simulated, and of course, for those, the input will be a pair of simulated particles, which will emerge as simulated entangled particles, as shown here. 
In real life however, a neural network may produce entanglement, particularly between time separated systems [26], [27], [28]. This can become a source of entanglement, and can be useful in understanding various computational capabilities of neural networks, particularly their amazing ability of resolving strongly correlated quantum systems in a way that is better than any algorithmic approach [29], [30].
It is probably not beyond doubt that the entanglement presented here is a result of the type of coupling introduced in an ad-hoc way in the differential equation. However, the actual equation for a real neural network, introduces much more coupling of the same type, linking the initial time for backpropagation to the integrated overall interaction hamiltonian for all considered times. Therefore, while additional structure may be expected, I am relatively confident this entanglement will not go away by introducing more realistic constructions. 

\section{Data Availability Statement}
Data sharing not applicable to this article as no datasets were generated or analysed during the current study.
\section{Competing interests}
There are no competing interests to declare
\section{Funding}
not applicable

\end{document}